\documentclass[11pt,a4paper]{article}
\usepackage[hyperref]{acl2021}
\usepackage{times}
\usepackage{latexsym}

\usepackage{xspace}
\usepackage{multirow}
\usepackage{hhline}
\usepackage{amsmath,amsfonts,amssymb}
\usepackage{pgfplots}
\pgfplotsset{compat=1.9}
\usepackage{graphicx}
\graphicspath{ {images/} }

\usepackage{float}
\restylefloat{table}

\usepackage{bm}
\usepackage{soul}
\usepackage[inline]{enumitem}
\usepackage{textcomp}
\usepackage{url}
\usepackage{xcolor}  %

\usepackage{booktabs, cellspace}
\usepackage{listings}%
\usepackage[textsize=tiny,disable]{todonotes}

\newcommand{\lm}{LM\xspace}
\newcommand{\mlm}{MLM\xspace}
\newcommand{\tm}{TM\xspace}

\newcommand{\bleu}{\textsc{bleu}\xspace}
\newcommand{\nmt}{NMT\xspace}

\newcommand{\ppl}{\textsc{ppl}\xspace}

\newcommand{\tto}{$\rightarrow$}
\newcommand{\tbi}{$\leftrightarrow$}

\newcommand{\shortpar}[1]{\vspace{3pt}\noindent\textbf{#1}~}

\setuldepth{0.0}
\newcommand{\trans}[2]{\textbf{#1}\tto{\textbf{#2}}}

\newcommand{\cell}[2]{$\text{#1}\color{gray}\scriptscriptstyle\pm\text{#2}$}
\newcommand{\best}[2]{$\textbf{#1}\color{gray}\scriptscriptstyle\pm\text{#2}$}
\newcommand{\gain}[2]{$\text{\ul{#1}}\color{gray}\scriptscriptstyle\pm\text{#2}$}

\newcommand{\scell}[2]{$\text{#1}$}
\newcommand{\sbest}[2]{$\textbf{#1}$}
\newcommand{\sgain}[2]{$\text{\ul{#1}}$}

\usepackage{lettrine}

\usepackage{microtype}

\aclfinalcopy %
\def\aclpaperid{2353} %

\title{Exploring Unsupervised Pretraining Objectives for Machine Translation}
\author{Christos Baziotis, Ivan Titov, Alexandra Birch \and Barry Haddow\\
    Institute for Language, Cognition and Computation \\
    School of Informatics, University of Edinburgh\\
    10 Crichton Street, Edinburgh EH8 9AB \\
  \texttt{c.baziotis@ed.ac.uk}, \texttt{ititov@inf.ed.ac.uk}, 
  \\
  \texttt{a.birch@ed.ac.uk}, \texttt{bhaddow@inf.ed.ac.uk}}
\date{}

\begin{document}
\maketitle
\begin{abstract}

Unsupervised cross-lingual pretraining has achieved strong results in neural machine translation (NMT), 
by drastically reducing the need for large parallel data.
Most approaches adapt masked-language modeling (MLM) to sequence-to-sequence architectures,
by masking parts of the input and reconstructing them in the decoder.
In this work, 
we systematically compare masking with alternative objectives
that produce inputs resembling real (full) sentences,
by reordering and replacing words based on their context.
We pretrain models with different methods on English$\leftrightarrow$German, English$\leftrightarrow$Nepali and English$\leftrightarrow$Sinhala monolingual data, and evaluate them on NMT.
In (semi-) supervised NMT, 
varying the pretraining objective leads to surprisingly small differences in the finetuned performance, 
whereas unsupervised NMT is much more sensitive to it.
To understand these results, 
we thoroughly study the pretrained models using a series of probes and verify that they encode and use information in different ways.
We conclude that finetuning on \textit{parallel} data is mostly sensitive to few properties that are shared by most models, such as a strong decoder, 
in contrast to unsupervised NMT that also requires models with strong cross-lingual abilities.
\end{abstract}

\setlist[itemize]{leftmargin=1em}

\setlength{\textfloatsep}{10pt plus 2pt minus 4pt}
\setlength{\dbltextfloatsep}{\baselineskip}
\setlength{\abovecaptionskip}{5pt}

\setlength{\abovedisplayskip}{5pt}
\setlength{\belowdisplayskip}{5pt}

\section{Introduction}
\todo{christos: Please resolve the comments that you think are addressed by the updated version.}
Neural machine translation (NMT) is notoriously data-hungry~\cite{koehn-knowles-2017-six}. 
To learn a strong model it requires large, high-quality and in-domain parallel data,
which exist only for a few language-pairs.
The most successful approach for improving low-resource NMT is backtranslation~\cite{sennrich-etal-2016-improving}, that exploits abundant monolingual corpora to augment the parallel with synthetic data.
However, in low-resource settings, it may fail to improve or even degrade translation quality
if the initial model is not strong enough
~\cite{imankulova-etal-2017-improving, burlot-yvon-2018-using}.

Unsupervised pretraining is a complementary technique, 
that has revolutionized many natural language understanding (NLU) tasks~\cite{wang2018glue}.
The dominant approach is to train a (large) 
model on a lot of unlabeled data using the masked language modeling (MLM;~\citet{devlin-etal-2019-bert}) objective
and then finetune it on a downstream task.
Besides improving generalization, 
good initialization drastically reduces the need for labelled data.
This paradigm has been applied recently to NMT yielding impressive results in low-resource settings,
with models such as XLM~\cite{Conneau2019-fk}, MASS~\cite{Song2019-dc} and BART/mBART~\cite{Lewis2019-ns,Liu2020-un},
that adapt MLM to sequence-to-sequence architectures.
Although pretraining alone is not enough to outperform backtranslation, 
it helps the initial model to produce synthetic data of sufficient quality,
and combining them yields further improvements.

\begin{figure}[t]
    \centering
    \includegraphics[width=0.97\columnwidth, page=7]{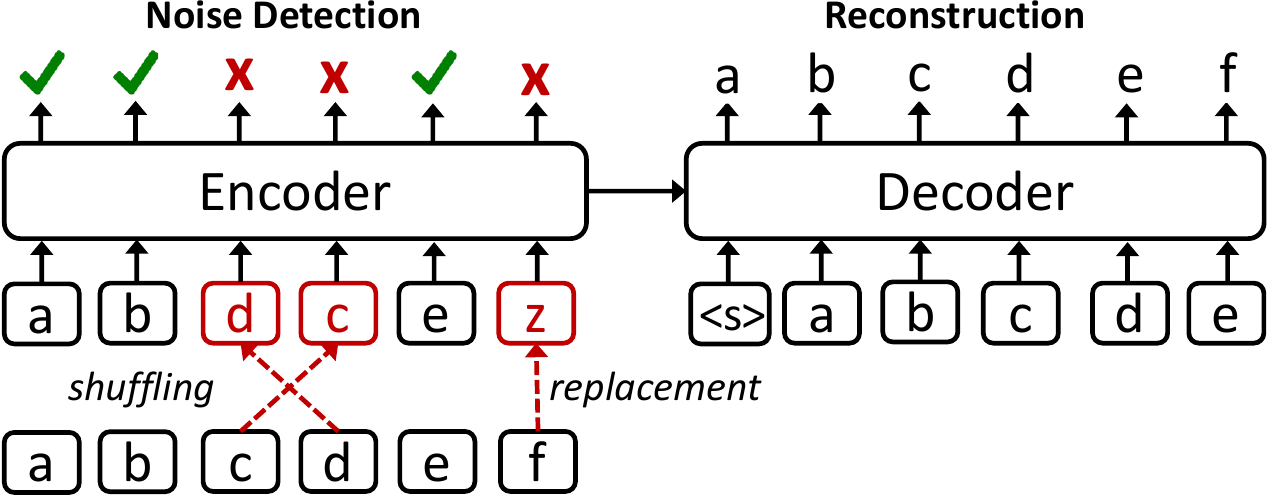}
    \caption{We consider noising methods that produce inputs which resemble real sentences, unlike masking.}
    \label{fig:seq2seq_pretrain}
    
\end{figure}

Most prior work in pretraining has focused on optimizing the masking strategy~\cite{TACL2257}.
Similarly, MASS and mBART consider slightly different masking strategies.
However, due to differences in their experimental setup (i.e., capacity or training data) and lack of analysis that goes beyond evaluation on downstream tasks,
it is unclear if there is a meaningful difference between them, 
as far as NMT is concerned.
They also suffer from a pretraining-finetuning discrepancy~\cite{yang2019xlnet},
in which a model is pretrained on masked inputs, but finetuned on full sentences.

In this work\footnotemark, we explore different objectives to masking for unsupervised cross-lingual pretraining.
We inject noise that creates examples (Fig.~\ref{fig:seq2seq_pretrain}), similar to those encountered in finetuning,
unlike masking.
This includes, 
randomly replacing input words based on their context
using a cross-lingual generator, inspired by~\citet{Clark2019-hu},
and locally reordering input words,
which prevents the cross-attention from naively (monotonically) attending over the source.
We also explore auxiliary losses over the encoder to improve its representations.

First, we pretrain models with different configurations, on English-German, English-Nepali and English-Sinhala monolingual data.
Then, we \textit{systematically} compare them on the downstream tasks of supervised, 
semi-supervised and unsupervised NMT.
In (semi-) supervised NMT, 
we observe that models yield surprisingly similar results,
although some methods are better than others.
We find that even pretraining with shuffled inputs leads to significant improvements over random initialization, 
similar to the concurrent work of \citet{Sinha2021MaskedLM} on pretrained encoders for NLU.
Unsupervised NMT, however, reveals large (up to 9 \bleu points) differences,
and against our expectations, masking achieves the best performance.
To understand these results, unlike prior work,
we thoroughly analyze the pretrained models using a series of probes,
and discover that each objective drives the models to encode and use information in unique ways.

Based on our findings,
we conclude that each finetuning process 
is sensitive to specific properties of pretrained models, 
similar to \citet{Artetxe_2020}.
We hypothesize that (semi-) supervised NMT 
is mostly sensitive to
the \lm abilities of pretrained models,
as the source$\rightarrow$target mappings can be learnt from the parallel data.
Unsupervised NMT requires models to also rely on their own word-translation abilities.
Our contributions are:
\setlist[enumerate]{leftmargin=20pt}
\begin{enumerate}
[topsep=3pt,itemsep=4pt,partopsep=0pt, parsep=0pt]

\item 
We \textit{systematically} compare many pretraining methods, including alternatives to masking, in three NMT tasks and for three language-pairs. 

\item We discover that (semi-) supervised NMT is not sensitive to the pretraining strategies. 
Our ablation (\S\ref{sec:ablation}) suggests that a strong decoder is the most important factor, 
while differences in the encoder (\S\ref{sec:xss}) don't affect the results.

\item Unsupervised setting is much more sensitive to the pretraining objective, and masking methods are the most effective. 
We hypothesise that learning to copy is important here (\S\ref{sec:analysis-entropy}) as is cross-lingual encoding (\S\ref{sec:xss}).

\item 
We analyze the pretrained models with a series of probes (\S\ref{sec:analysis-identifiability}, \S\ref{sec:analysis-entropy}, \S\ref{sec:decoder-sensitivity}), 
and show noticeable differences in how they encode and use information,
offering valuable insights.

\end{enumerate}

\footnotetext{Code at \href{https://github.com/cbaziotis/nmt-pretraining-objectives}{github.com/cbaziotis/nmt-pretraining-objectives}}
\section{Related Work}

\noindent\textbf{Pretraining for \nmt}
\citet{ramachandran-etal-2017-unsupervised} first explored unsupervised pretraining for \nmt
using \lm{s} trained on monolingual data of the source and target languages to initialize the encoder and decoder of an RNN-based \tm~\cite{Bahdanau2014}.
\citet{Conneau2019-fk} adopt the same approach, by extending BERT/MLM~\cite{devlin-etal-2019-bert} to the cross-lingual setting (XLM). 
They randomly mask tokens from input sentences in many languages, 
and the model is trained to predict them.
However, the same pretrained XLM is used to (separately) initialize both the encoder and decoder of a downstream translation model (\tm), which neglects the interaction between them.
MASS~\cite{Song2019-dc} addresses this limitation, by extending MLM to sequence-to-sequence pretraining, which includes the cross-attention mechanism,
and achieved further improvements in low-resource and unsupervised \nmt.
\citet{Liu2020-un}, concurrently demonstrated comparable results with a similar approach (mBART), but on a larger scale.
Both mBART and MASS, consider different strategies for reconstructing masked input spans. 

\shortpar{Objectives}
\citet{yang2019xlnet} point out that BERT~\cite{devlin-etal-2019-bert} is pretrained with masked inputs, but then finetuned on full sentences, which creates a discrepancy.
To address this, they change the self-attention in Transformers~\cite{vaswani2017attention} to predict tokens conditioned on all permutations of other tokens in a sentence and
\citet{Song2020-ps} extend this to sequence-level pretraining for NLU.
MARGE~\cite{Lewis2020-uh} explores multi-lingual pretraining for document-level NMT, 
by reconstructing texts from a set of retrieved relevant documents.
\citet{Clark2019-hu} propose the replaced token detection (RTD) objective for pretraining text encoders.
They replace tokens with samples from a \mlm and train the encoder as a discriminator to predict whether each word is real or fake. 
Similar ideas have been previously explored in \nmt with contextual data augmentation~\cite{Fadaee2017-fv,Kobayashi2018-dc, Gao2019-zs}.

\section{Pretraining}
Our pretraining model is a multilingual denoising sequence autoencoder, based on the Transformer~\citep{vaswani2017attention}.

We assume access to a corpus of unpaired data, containing text in two languages $A$, $B$. 
Given a text sequence of $N$ tokens $\bm{x}= \langle x_1, x_2, ..., x_N \rangle$ we first add noise to it and obtain its corrupted version $\bm{x'}$.
An encoder transforms $\bm{x'}$ into a sequence of contextualized representations $h(\bm{x'})=\langle h_1, h_2, ..., h_N\rangle$, 
which are given as input to the decoder, that produces a reconstruction of $\bm{x}$.
The reconstruction loss is the negative log-likelihood (NLL)
of $\bm{x}$:
\begin{align}
    \mathcal{L_\textsc{R}} = \frac{1}{N} \sum_{t=1}^{N} - \log p (\bm{x}_t|\bm{x}_{<t}, h(\bm{x'}))
\end{align}
\noindent 
Each batch contains sentences in either $A$, or $B$ and to distinguish between them, 
we add language id tokens at the end of the source sentences, 
and the beginning of the target sentences.

\subsection{Pretraining Methods}
In this section, we describe the methods that we use to inject noise into the model.
We also explore auxiliary losses over the encoder, aiming to improve the input representations.

\shortpar{Masking}
Similar to prior work, we replace a random subsequence $M$ of the input tokens with a special \texttt{[MASK]} token and train the model to reconstruct the original input. We consider masking words as well as spans following mBART.  

\shortpar{Masking + eMLM} When using masking noise we also explore
the addition of an auxiliary MLM loss over the \textit{encoder} to which we will refer as eMLM.
This explicitly trains the encoder to reconstruct the representations of masked tokens:
\begin{align}
    \mathcal{L_{\textrm{e}\textsc{mlm}}} = \frac{1}{\lvert M\rvert} 
                               \sum_{t \in M} - \log p (\bm{x}_t|\bm{x}_{\not\in M}))
\end{align}

\shortpar{Replacing}
We inject word replacement noise, by extending~\citet{Clark2019-hu} to the cross-lingual setting.
Specifically, we jointly train a \textit{separate} cross-lingual (mBERT-like) MLM generator.
First, we mask a random subset $M$ of the input tokens\footnotemark and the generator is trained to predict them:
\begin{align}
    \mathcal{L_\textsc{g}} = \frac{1}{\lvert M\rvert} 
                               \sum_{t \in M} - \log p (\bm{x}_t|\bm{x}_{\not\in M}))
\end{align}
For each masked token $\bm{x}_t$, 
the generator produces a distribution 
$p_\textsc{g}(\bm{x}_t|\bm{x}_{\ne t})$. 
We replace the masked tokens with samples from $p_\textsc{g}$ to obtain $\bm{x'}$, and feed the updated (corrupted) input to the encoder.
\footnotetext{We use whole-word masking, that masks all the (subword) tokens of a word, instead of independent token masking.}
We consider two configurations for the generator:
\begin{itemize}[topsep=0pt,itemsep=2pt,partopsep=0pt, parsep=0pt]
\item 
\textit{Untied}: The default setting, we use a small generator, which is half the size (x0.5 parameters) of the encoder, following~\citet{Clark2019-hu}.
\item 
\textit{Tied}: We tie the weights of the encoder and the generator.
This setting implicitly adds an auxiliary MLM loss over the encoder, and can be thought as a counterpart of ``replace+eMLM''.
\end{itemize}
\shortpar{Replacement + RTD}
Motivated by the results of~\citet{Clark2019-hu} in monolingual NLU,
we add a replacement token detection (RTD) head over the encoder
that gives direct supervision to the model regarding the location of noise.
The RTD head $D(\cdot)$ is a token-level discriminator over the encoder outputs $h(\bm{x'})$, 
that predicts if an input token $\bm{x'_t}$ is original or replaced.
We parameterize $D$ with a non-linear projection followed by a sigmoid function: 
$D(\bm{x'_t}) = \textrm{sigmoid}(\bm{u}^\top \textsc{relu}(\bm{W_D} \, h(\bm{x'_t})))$.
The RTD loss is defined as the average token-level binary cross-entropy:
\begin{align}
    \mathcal{L_\textsc{rtd}} =  \frac{1}{N}
    \sum_{t=1}^{N} &- (\bm{x'}_t=\textrm{original}) \log D(\bm{x'}_t) \\
                   &- (\bm{x'}_t\neq\textrm{original}) (1-D(\bm{x'}_t)) \nonumber
\end{align}
\shortpar{Shuffling}
Although pretrained models, such as MASS or mBART, do pretrain the cross-attention mechanism, the input words remain in their original positions
and this biases the models into learning only naive monotonic alignments.
To \textit{actively} pretrain the cross-attention, we locally shuffle a random subset of \textit{whole-words} in the input, using the method of~\cite{lample2018unsupervised}.
The length of reordering is bounded by $k$ (positions).
Small $k$ introduce local shuffling, while large $k$ allow words to be moved farther from their original position, making the input more like a bag-of-words (BoW).

\subsection{Optimization}
During pretraining, we minimize a weighted sum of the reconstruction $L_\textsc{R}$, and depending on the method, some of the auxiliary $L_{\textrm{e}\textsc{mlm}}$, $L_\textsc{g}$, $L_\textsc{rtd}$ losses.
We assign equal weight ($\lambda=1$) to all losses, 
except for $L_\textsc{rtd}$ for which we set its weight $\lambda=25$ 
following \citet{Clark2019-hu}, to account for the fact that is in a different scale.

\begingroup
\setlength{\tabcolsep}{9.0pt} %
\renewcommand{\arraystretch}{1.0} %
\begin{table*}[!tb]
	\small
	\centering
	\begin{tabular}{lrrrrrrrr}
		\toprule
		\textbf{Method} & \multicolumn{2}{c}{\trans{en}{de}} & \multicolumn{2}{c}{\trans{de}{en}}
		& \trans{en}{ne} & \trans{ne}{en} & \trans{en}{si} & \trans{si}{en} \\ \cmidrule(lr){2-3} \cmidrule(lr){4-5}
		                              & \multicolumn{1}{r}{wmt18} & \multicolumn{1}{r}{wmt19} & wmt18 & wmt19 &     &      &     &      \\ \midrule
		random                      & \cell{26.2}{0.1}   & \cell{25.3}{0.1} & \cell{27.6}{0.1}  & \cell{19.1}{0.3}  		& \cell{3.3}{0.1} & \cell{6.5}{0.1} 		& \cell{2.5}{0.1} & \cell{6.5}{0.1} \\
		mask=35\%                     & \cell{33.3}{0.1}   & \cell{30.7}{0.2} & \cell{33.2}{0.0}  & \best{25.4}{0.0}  		& \cell{5.1}{0.1} & \cell{10.2}{0.1} 		& \cell{3.7}{0.0} & \cell{10.0}{0.1} \\
		mask=35\% +eMLM                & \cell{33.4}{0.0}   & \cell{30.6}{0.1} & \cell{33.5}{0.1}  & \cell{25.2}{0.2}  		& \best{5.3}{0.0} & \best{10.8}{0.1} 		& \best{4.0}{0.0} & \cell{10.4}{0.1} \\
		mask=35\% (span)              & \cell{33.3}{0.1}   & \cell{30.5}{0.1} & \cell{33.4}{0.0}  & \cell{25.2}{0.0}  		& \cell{5.1}{0.1} & \cell{10.1}{0.1} 		& \cell{3.9}{0.1} & \cell{9.9}{0.1} \\
		shuffle=5                     & \cell{31.6}{0.1}   & \cell{28.7}{0.0} & \cell{31.7}{0.0}  & \cell{23.9}{0.1}  		& \cell{4.9}{0.0} & \cell{9.9}{0.1} 		& \cell{3.4}{0.0} & \cell{10.1}{0.1} \\
		replace=35\%                  & \cell{33.9}{0.0}   & \cell{30.3}{0.2} & \cell{33.5}{0.1}  & \best{25.4}{0.1}  		& \cell{5.1}{0.1} & \cell{9.9}{0.1} 		& \cell{3.7}{0.0} & \cell{9.8}{0.0} \\
		replace=35\% +RTD             & \cell{32.9}{0.1}   & \cell{30.0}{0.0} & \cell{32.5}{0.0}  & \cell{24.4}{0.1}  		& \cell{5.0}{0.0} & \cell{9.9}{0.1} 		& \cell{3.4}{0.1} & \cell{9.7}{0.2} \\
		replace=35\% +tied            & \best{34.2}{0.0}   & \best{30.8}{0.1} & \best{33.7}{0.1}  & \cell{25.3}{0.2}  		& \best{5.3}{0.0} & \cell{10.6}{0.1} 		& \cell{3.7}{0.0} & \best{10.5}{0.1} \\ \cmidrule{2-9}
		\quad\quad+ shuffle=3         & \cell{34.0}{0.0}   & \gain{31.1}{0.1} & \cell{33.4}{0.1}  & \cell{25.1}{0.2}  		& \gain{5.5}{0.0} & \gain{11.0}{0.0} 		& \gain{4.0}{0.0} & \gain{10.8}{0.1} \\
		\bottomrule
	\end{tabular}
	\caption{Supervised NMT results.
		We report the average of 3 runs and the standard error of the mean (SEM).}
	\label{table:sup_nmt}
\end{table*}
\endgroup 
\section{Experiments}\label{sec:experiments}
\shortpar{Datasets}
We focus on low-resource translation and consider three diverse language-pairs: English-German, English-Nepali and English-Sinhala. 
For English-German, we use the low-resource WMT News Commentary v13~\cite{bojar-etal-2018-findings}~\footnote{\href{http://www.statmt.org/wmt18/translation-task.html}{http://www.statmt.org/wmt18/translation-task.html}} parallel dataset, which contains approximately 275K sentences. 
For pretraining, we use as monolingual data the
WMT News Crawl articles~\cite{bojar-etal-2018-findings} from the year 2007 to 2017, which comprise 190M and 270M sentences for English and German, respectively.
For English-Nepali and English-Sinhala, we use the same data as in~\citet{guzman-etal-2019-flores}. 
The (pretraining) monolingual data contain 5M sentences from Common Crawl and Wikipedia per language, 
while the parallel data are approximately 600K sentences from the Bible, Open Subtitles, GNOME/KDE/Ubuntu, and Paracrawl.

\shortpar{Pre-processing}
For Nepali and Sinhala, we use the preprocessing scripts\footnotemark~provided by~\citet{guzman-etal-2019-flores}, whereas
for English and German, we use directly the raw data without any preprocessing.
We use sentencepiece~(SPM; \citet{kudo-richardson-2018-sentencepiece}) with the ``unigram'' model,
to train a subword-unit tokenization model on the concatenation of the monolingual data of each language-pair.
We learn a joint vocabulary of 60K symbols for the English-German models, 
and 20K symbols for the English-Nepali and English-Sinhala models.

\footnotetext{\href{https://github.com/facebookresearch/flores}{https://github.com/facebookresearch/flores}}

\shortpar{Evaluation}
For English-German, we use the WMT \textit{newstest2017} as dev-set and the \textit{newstest2018} and \textit{newstest2019} as test-sets.
For English-Nepali and English-Sinhala, we use the evaluation datasets provided by~\citet{guzman-etal-2019-flores}, which are drawn from Wikipedia articles.
We evaluate models using \bleu~\cite{papineni2002bleu} computed with Sacre\textsc{bleu}~\cite{post-2018-call}. 
We report detokenized \bleu when translating into German and English,
and tokenized \bleu when translating into Nepali and Sinhala, following~\citet{guzman-etal-2019-flores}. 
At test time, we decode with beam search, using beams of size 5.

\shortpar{Model and Training}
Our models are based on the Transformer architecture~\cite{vaswani2017attention}.
We use the Transformer-base configuration to reduce the computational cost and be able to explore more methods. 
We describe in detail the model architecture and hyperparameters, as well as the pretraining and finetuning processes, in Appendix~\S\ref{sec:model-config}.
Our code is based on the official mBART implementation in Fairseq~\cite{ott2019fairseq}.

\subsection{Supervised Translation}\label{sec:sup-nmt}
In our first experiment (Table~\ref{table:sup_nmt}), we evaluate each pretrained model on supervised \nmt by finetuning it on the parallel data. 
As a baseline we use a randomly initialized \tm with identical configuration and vocabulary to that of the pretrained models, denoted as ``random''.
We also pretrain a model equivalent to mBART denoted as ``mask (span)''.

\shortpar{Results}
All pretraining methods yield large improvements over random initialization, in all directions and language pairs.
The differences between each method are more pronounced in en\tbi{de},
whereas in en\tbi{ne} and en\tbi{si}, all models reach much lower scores, especially in en\tto{X}, probably because of low-quality training data, and a domain mismatch between the parallel and test data.

When we compare each type of input noise in isolation, 
we observe that masking and replacement achieve similar results,
and both are better than shuffling.
However, pretraining with shuffling noise alone still yields surprisingly strong results.
It improves over random initialization in all experiments,
and it even reaches similar \bleu scores to masking and replacements in en\tbi{ne} and en\tbi{si}.
Note that, the common denominator in all pretraining methods is the decoder, 
which is trained as a conditional LM with teacher forcing 
and the only difference is the input conditioning context.
This implies that \textit{one} key factor in pretraining for NMT is improving the LM capabilities of the decoder.

\shortpar{Auxiliary Losses}
The best results are obtained by ``mask+eMLM'' and ``replace+tied'', both of which benefit from an encoder MLM loss.
We observe, that eMLM and tying are more effective in the X\tto{en} than en\tto{X} direction, especially for en\tbi{ne} and en\tbi{si}.
This makes intuitive sense because eMLM improves the representations of the encoder,
which is more important for languages with limited or low-quality data.
Specifically, 
we observe that adding eMLM improves \bleu by +0.7 for ne\tto{en}, +0.4 for si\tto{en}
and tying the generator with the encoder yields +0.7 \bleu for ne\tto{en}, +0.7  \bleu for si\tto{en}.
\citet{Wang2020-nd} make a similar observation in experiments in multilingual NMT.

Incorporating RTD, however, has a negative effect in most experiments.
This is unexpected, given that~\citet{Clark2019-hu} showed that pretraining text encoders with RTD outperformed MLM in NLU tasks.
This warns us that methods which produce strong encoders for NLU might not necessarily improve encoders for NMT.
Note that, 
\citet{siddhant2020evaluating, Wang2020-nd} have discovered similar surprising results in cross-lingual NLU tasks.

\shortpar{Noise Combination}
We also consider a combination of the best replacement-based variant ``replace+tied'' with shuffling.
Shuffling is applied after the replacements are sampled and we limit the length of reordering to $k=3$ 
to prevent extreme corruption of the input.
We observe that this combination yields small gains in most experiments.
We also explored more pretraining methods and configurations, including the injection of noise into the decoder, but they didn't produce significant differences in terms of \bleu (see Appendix~\ref{sec:decoder-noise}).

\begingroup
\setlength{\tabcolsep}{3.5pt} %
\renewcommand{\arraystretch}{0.97} %
\begin{table}[!tb]
	\small
	\centering
	\begin{tabular}{@{}lrrrr@{}}
		\toprule
		\textbf{Method} & \multicolumn{2}{c}{\trans{en}{de}} & \multicolumn{2}{c}{\trans{de}{en}} \\
		\cmidrule(lr){2-3} \cmidrule(lr){4-5}
		                      & wmt18 & wmt19 & wmt18 & wmt19 \\\midrule
		mask=15\%             & \cell{32.7}{0.1}  & \cell{30.4}{0.1}  & \cell{32.9}{0.0}  & \cell{25.1}{0.1}  \\
		mask=35\%             & \best{33.3}{0.1}  & \best{30.7}{0.2}  & \best{33.2}{0.0}  & \best{25.4}{0.0}  \\
		mask=50\%             & \cell{33.2}{0.0}  & \cell{30.4}{0.0}  & \cell{33.1}{0.1}  & \cell{25.2}{0.1}  \\ \midrule
		shuffle=3             & \cell{30.3}{0.1}  & \cell{27.9}{0.0}  & \cell{30.5}{0.0}  & \cell{22.9}{0.1}  \\
		shuffle=5             & \best{31.6}{0.1}  & \best{28.7}{0.0}  & \best{31.7}{0.0}  & \best{23.9}{0.1}  \\ \midrule
		replace=15\%          & \cell{33.8}{0.1}  & \best{30.4}{0.1}  & \cell{33.1}{0.1}  & \cell{25.3}{0.2}  \\
		replace=35\%          & \best{33.9}{0.0}  & \cell{30.3}{0.2}  & \best{33.5}{0.1}  & \best{25.4}{0.1}  \\
		replace=50\%          & \cell{33.3}{0.0}  & \cell{30.3}{0.1}  & \cell{32.8}{0.0}  & \cell{24.7}{0.2}  \\ \midrule
		replace=35\% (1.0x)   & \cell{33.7}{0.1}  & \cell{30.6}{0.0}  & \cell{33.3}{0.0}  & \cell{25.0}{0.1}  \\
		replace=35\% +nucleus & \best{33.9}{0.0}  & \best{30.9}{0.1}  & \best{33.6}{0.0}  & \best{25.3}{0.0}  \\ \midrule
		replace=35\% +RTD=4   & \best{33.2}{0.0}  & \cell{29.8}{0.1}  & \cell{32.4}{0.1}  & \best{24.5}{0.0}  \\
		replace=35\% +RTD=6   & \cell{32.9}{0.1}  & \best{30.0}{0.0}  & \best{32.5}{0.0}  & \cell{24.4}{0.1}  \\
		\bottomrule
	\end{tabular}
	\caption{Supervised NMT results (mean and SEM of 3 runs), for different configurations of pretrained models.}
	\label{table:sup_nmt_comp}
\end{table}
\endgroup

\subsubsection{Parameter Sensitivity Analysis}\label{sec:sup-sensitivity-analysis}
We also explore how changing key parameters of each pretraining method affects performance in supervised NMT.
We report the results in Table~\ref{table:sup_nmt_comp}.
We observe there is a ``sweet-spot'' for the amount of noise used in each method.
Shuffling shows larger variability 
and we find that by making the token swaps less local (i.e., increasing $k$ that makes the input more BoW),
yields better results. 

\shortpar{Generator Size}
Next, we focus on why tying the encoder and generator yields better results. 
Either the encoder benefits from the implicit MLM loss, 
or tying improves the generator and consequently its samples.
We train an \textit{untied} model with equal capacity to the encoder ``replace=35\% (x1.0)'', and a similar model, but we sample replacements with nucleus sampling\footnotemark~\cite{holtzman2020the} with  top-p=$0.9$, to avoid low-probability tokens.
Neither of those variants yields any measurable difference with ``replace=35\%'', which suggests that MLM is responsible for the improvements.

\shortpar{RTD Position}
We also explore if the position of the RTD head is responsible for its negative effects in NMT,
as it might force the encoder to preserve information irrelevant for NMT to its outputs.
To test this, we train a model with RTD over its fourth (RTD=4) instead of last/top (RTD=6) layer. 
We find that this change has a marginal effect on \bleu.

\begingroup
\setlength{\tabcolsep}{9.5pt} %
\renewcommand{\arraystretch}{1} %
\begin{table*}[!t]
	\small
	\centering
	\begin{tabular}{lllrrrrrr}
		\toprule
		\textbf{Method} & \multicolumn{2}{c}{\trans{en}{de}} & \multicolumn{2}{c}{\trans{de}{en}} 
		& \trans{en}{ne} & \trans{ne}{en} & \trans{en}{si} & \trans{si}{en} \\ \cmidrule(lr){2-3} \cmidrule(lr){4-5}
		                             & \multicolumn{1}{r}{wmt18} & \multicolumn{1}{r}{wmt19} & wmt18 & wmt19 &     &      &     &      \\ \midrule
		random                       & \scell{34.8}{} & \scell{29.2}{} & \scell{37.4}{}  & \scell{24.8}{}  & \scell{5.8}{} & \scell{13.9}{} & \scell{6.5}{} & \scell{12.9}{} \\
		mask=35\%                    & \scell{38.3}{} & \scell{31.5}{} & \scell{38.8}{}  & \scell{27.7}{}  & \scell{6.3}{} & \scell{14.9}{} & \scell{6.5}{} & \scell{13.5}{} \\
		mask=35\% +eMLM              & \scell{38.4}{} & \sbest{31.8}{} & \sbest{39.0}{}  & \scell{27.4}{}  & \sbest{6.6}{} & \scell{15.0}{} & \sbest{7.3}{} & \sbest{14.2}{} \\
		mask=35\% (span)             & \scell{38.3}{} & \scell{31.6}{} & \scell{39.0}{}  & \scell{27.5}{}  & \scell{6.4}{} & \scell{14.4}{} & \scell{6.3}{} & \scell{13.4}{} \\
		shuffle=5                    & \scell{36.2}{} & \scell{30.8}{} & \scell{36.9}{}  & \scell{26.1}{}  & \scell{6.4}{} & \scell{13.5}{} & \scell{6.3}{} & \scell{11.9}{} \\
		replace=35\%                 & \scell{38.2}{} & \scell{31.2}{} & \scell{38.6}{}  & \scell{27.5}{}  & \scell{6.4}{} & \scell{14.7}{} & \scell{6.0}{} & \scell{13.1}{} \\
		replace=35\% +RTD            & \scell{37.7}{} & \scell{31.3}{} & \scell{38.2}{}  & \scell{27.5}{}  & \scell{6.1}{} & \scell{14.2}{} & \scell{5.9}{} & \scell{12.8}{} \\
		replace=35\% +tied           & \sbest{38.5}{} & \scell{31.7}{} & \scell{38.8}{}  & \sbest{27.8}{}  & \scell{6.4}{} & \sbest{15.1}{} & \scell{6.6}{} & \scell{13.7}{} \\  \cmidrule{2-9}
		\quad\quad+shuffle=3         & \scell{38.2}{} & \scell{31.6}{} & \scell{38.6}{}  & \scell{27.0}{}  & \scell{6.5}{} & \sgain{15.4}{} & \sgain{7.4}{} & \scell{14.2}{} \\
		\bottomrule
	\end{tabular}
	\caption{Finetuning results to \textbf{semisupervised} NMT.
	Each \tm is trained on the concatenation of real and backtranslated sentences, 
	obtained by a backward \tm initialized from the same pretrained model. 
	}
	\label{table:semi_nmt}
\end{table*}
\endgroup
\raggedbottom

\subsection{Semi-supervised Translation}\label{sec:semi-nmt}
In Table~\ref{table:semi_nmt} we report results for semi-supervised NMT,
using backtranslation.
Both the forward and backward \tm are initialized from the same model.
We generate the backtranslations from the same monolingual data that we used for pretraining with greedy-sampling, which is preferable for weak \tm{s}~\cite{edunov-etal-2018-understanding}, and upsample the real data to maintain a 1:1 ratio with the synthetic data.

\shortpar{Results}
Backtranslation yields significant gains in all experiments, 
and the initialization from pretrained models boosts performance even more.
The relevant performance between methods 
is consistent with the supervised NMT, but their differences shrink further.
We suspect that adding more data narrows the room for improvement of pretraining.

One exception is the initialization from ``shuffle=5'', 
which yields marginal gains in en\tbi{de} 
and even fails to reach the randomly initialized model in si\tbi{en} and ne\tbi{en}.
Note that, one of the advantages of backtranslation is improving the LM capabilities of the decoder with the addition of more clean target data.
We hypothesize that shuffling noise mainly pretrains the decoder as a LM,
and its benefits are largely neutralized by backtranslation.

\begingroup
\setlength{\tabcolsep}{4.5pt} %
\renewcommand{\arraystretch}{1} %
\begin{table}[!tb]
	\small
	\begin{tabular}{lrrrr}
		\toprule
		\textbf{Method}          & \multicolumn{2}{c}{\trans{en}{de}} & \multicolumn{2}{c}{\trans{de}{en}} \\
		\cmidrule(lr){2-3} \cmidrule(lr){4-5}
		                        & wmt18 & wmt19 & wmt18 & wmt19 \\\midrule
		mask=35\%               & \textbf{25.6}  & \textbf{19.1}  & \textbf{29.3}  & \textbf{20.3}  \\
		mask=35\% +eMLM         & 25.2  & 18.7  & 28.9  & 19.8  \\
		mask=35\% (span)        & 24.7 & 18.1  & 28.3  & 19.7  \\
		shuffle=5               & 17.1  & 13.1  & 20.5  & 15.6  \\
		replace=35\%            & 23.3  & 17.5  & 27.4  & 19.4  \\
		replace=35\% +RTD       & 22.8  & 16.8  & 26.5  & 18.5  \\
		replace=35\% +tied      & 24.1  & 17.8  & 27.8  & 19.3  \\ 
		\bottomrule
	\end{tabular}
	\caption{Finetuning results to \textbf{unsupervised} NMT. 
	}
	\label{table:unmt_main}
\end{table}
\endgroup

\subsection{Unsupervised Translation}\label{sec:unsup-nmt}
In Table~\ref{table:unmt_main} we evaluate the pretrained models on unsupervised NMT~\cite{artetxe2018unsupervised, lample2018unsupervised}, using \textit{only} the monolingual data.
In this experiment we focus on en\tbi{de}, because unsupervised NMT in en\tbi{ne} and en\tbi{si} yields very low \bleu scores~\cite{guzman-etal-2019-flores,Liu2020-un}.
In each batch, we generate backtranslations on-the-fly in the target language and the model is optimized to reconstruct the original sentences (i.e., $\mathbf{\textrm{en}}{\rightarrow}\mathbf{\textrm{de}}'{\rightarrow}\hat{\mathbf{\textrm{en}}}$),
following the same finetuning process as mBART (see Appendix~\ref{sec:appendix-unmt} for details).

\shortpar{Results} 
Unlike the experiments on parallel data (\S\ref{sec:sup-nmt}, \S\ref{sec:semi-nmt}), 
unsupervised NMT reveals large differences between pretrained models.
Strikingly, ``shuffle=5'' yields the lowest \bleu scores by a large margin.
This further supports the hypothesis that its primary strength is its decoder.
We hypothesize that all pretraining methods produce strong decoders, but not encoders.
Since in supervised NMT
the models can learn source-to-target mappings from the parallel data,
models with better cross-lingual abilities have a small edge.
However, in the unsupervised setting having a fluent decoder alone is not enough
as the models have to rely on their own word-translation capabilities~(\S\ref{sec:xss}) to be able to produce sufficient backtranslations.

Another unexpected result is that pretraining with masked inputs outperforms replacements.
Replacement-based models should intuitively be favoured,
because the mistakes injected by the generator during pretraining resemble those produced by backtranslation.
Masking-based models, however, are exposed to very different inputs without any signal from parallel data to help them transition to the new training regime, unlike supervised~NMT.

In our analysis (\S\ref{sec:analysis-entropy}), we find that masking biases models towards copying from the input, whereas replacements make models more ``cautious'' because some input words are fake.
We hypothesize that the ability of mask-based models to copy words, such as dates or named entities, is critical 
to kickstart the backtranslation process.

\subsection{Supervised Translation Ablations}\label{sec:ablation}
To estimate how important of each part of the pretrained model is for NMT, we conduct an ablation experiment.
First, we transfer all the weights of a pretrained model to a downstream model, 
except the weights of the ablated component.
Next, we freeze the pretrained weights for all components except for the ablated one, which we reinitialise randomly. Finally we   finetune only the ablated component,
to isolate the effects on the final score to the component 
and prevent the other components from compensating (details in Appendix~\ref{sec:appendix-ablation}).

\begin{figure}[t]
    \centering
    \includegraphics[width=1\columnwidth]{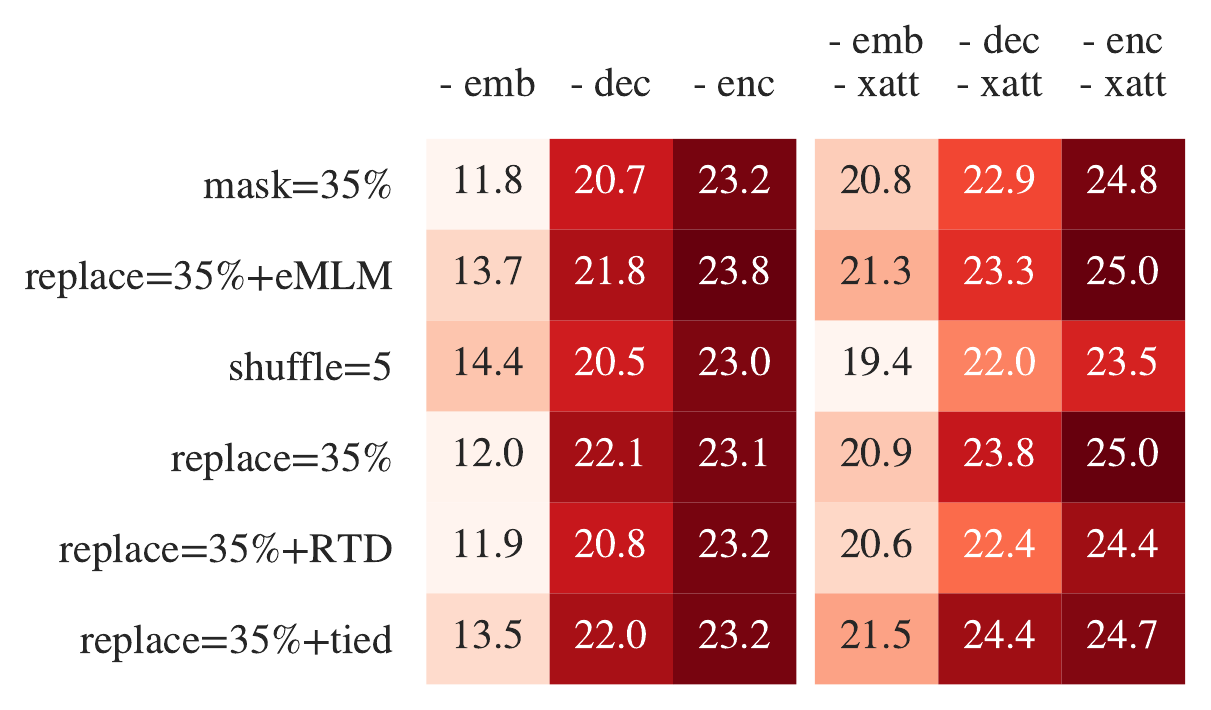}
    \caption{
    Ablation results for supervised NMT on de\tto{en} (wmt19). 
    We reset each main component individually (left) and with the cross-attention (right).
    }
    \label{fig:ablation-deen-merged-frozen}
\end{figure}

We divide the model into four parts: (1) the \textit{embedding} matrix, which is used for both the encoder and decoder embeddings and the vocabulary (output) projection, (2) the \textit{encoder} layers, 
(3) the \textit{decoder} layers and (4) the \textit{cross-attention} layers, which link the decoder with the encoder.
We report the de\tto{en} supervised NMT ablation results in Figure~\ref{fig:ablation-deen-merged-frozen}.
Higher \bleu scores show that a model can better recover after an ablation, 
which we interpret as an indication that the ablated parameters are less important.
We ablate each component both individually and combined with cross-attention, 
to allow the model to better learn to connect source and target representations.
We find in both settings, that
all models recover better after resetting their encoders than their decoders,
implying that the pretrained decoder parameters are more important.

\subsection{Cross-lingual Sentence Retrieval}\label{sec:xss}
\begin{figure}[t]
    \centering
    \includegraphics[width=1\columnwidth]{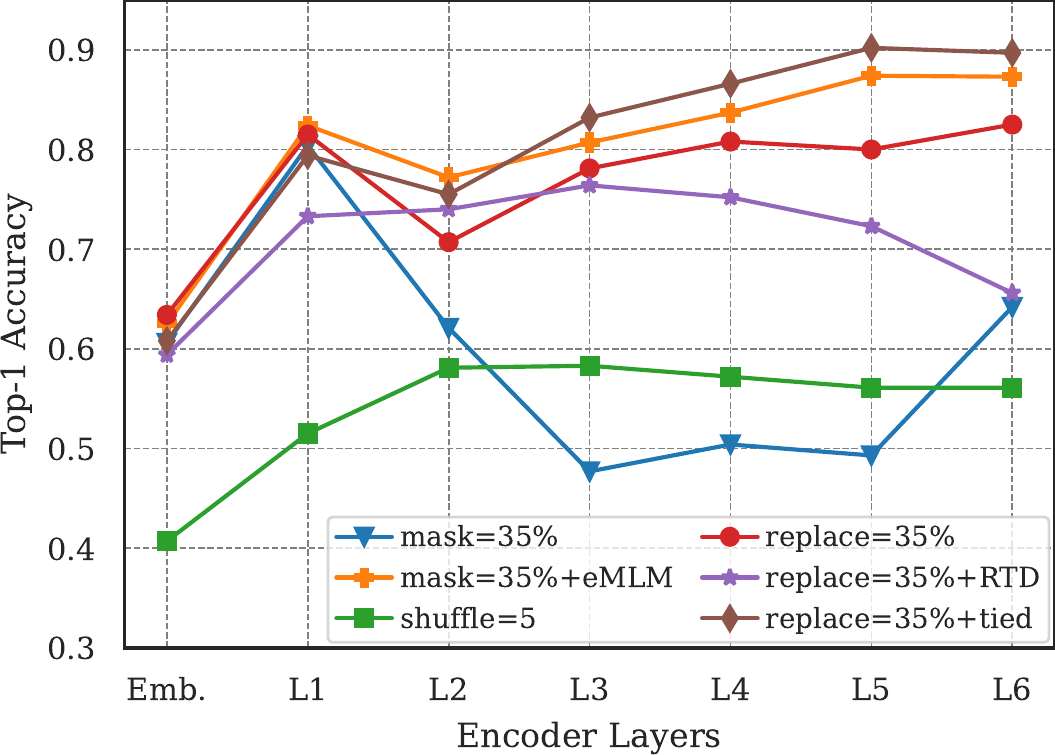}
    \caption{Parallel sentence retrieval accuracy (de\tto{en}). }
\label{fig:tatoeba_cosine_centered_selected_models}
\end{figure}

To study the cross-lingual abilities of the encoder of each pretrained model, 
we evaluate them on parallel sentence retrieval. 
In Figure~\ref{fig:tatoeba_cosine_centered_selected_models}, we report the (per layer) accuracy of each encoder, on the de\tto{en} Tatoeba test set~\cite{tiedemann-2020-tatoeba}\footnotemark.

The results indicate that different objectives do affect the encoder's cross-lingual abilities. 
The fact that this is not reflected in the \bleu scores of the finetuned models
further supports that even a small parallel data-set is  enough to align them to similar degrees. 
As hypothesized, shuffling induces the least effective cross-lingual representations. 
The RTD loss inhibits cross-linguality, as accuracy decreases for each layer closer to the RTD loss (L6).
The ``mask'' model exhibits an unexpected behaviour, 
where its accuracy drops in its middle layers before rising up again in its output layer.
We do not have a satisfying explanation for this and we leave it for future work.
However, adding an MLM loss over the encoder completely changes this behaviour,
and both ``mask+eMLM'' and ``replace+tied'' yield the best results.

We also notice an interesting result in the accuracy of the embeddings.
Shuffling noise induces poor alignment, 
unlike the other methods that have much better aligned embeddings.
We believe that this result is connected to the unsupervised NMT results.
Note that, the quality of embeddings affect the decoder as well, 
as their embeddings are tied.

\footnotetext{We follow~\citet{libovicky-etal-2020-language} 
and obtain the encoder sentence representations for each layer with mean pooling, 
followed by zero-centering per language (separately). Then, for each source sentence we retrieve its nearest neighbor from the target sentences based on cosine distance.}

\section{Analysis}\label{sec:analysis}
Finetuning on parallel data drives models towards similar destinations
(\bleu), 
but the results in unsupervised NMT hinted that their starting points are different, 
which implies that the finetuning process itself is critical.
We shift our focus to the pretrained models themselves 
and using a series of probes 
we study how pretraining methods affect their behaviour and the knowledge that they encode.
For clarity, here we discuss only the en\tbi{de} results. 
The en\tbi{ne} and en\tbi{si} results are included in Appendix~\ref{sec:appendix-analysis} and are consistent with this analysis.

\subsection{\hspace{-1pt}Encoder Denoising Capabilities}\label{sec:analysis-identifiability}
\todo{christos: I don't like the name ``Identifiability''. 
I used it because it was used by \citet{brunner2019identifiability}.
Do you have an alternative to propose? Something with reconstruction, denoising or recovery in it? Essential we measure how much of the denoising is happening in the encoder.}

\begin{figure}[t]
\centering
\includegraphics[width=1\columnwidth]{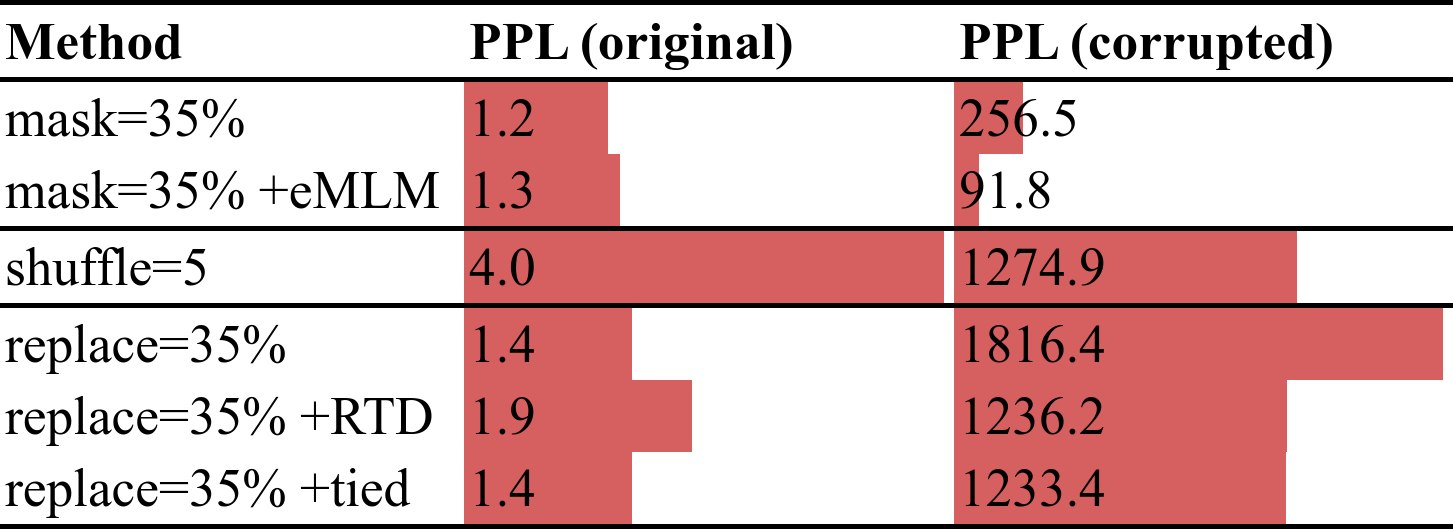}
\caption{Perplexity (\ppl~$\downarrow$) of the token prediction probe for the en\tbi{de} pretrained models.}
\label{fig:token_prediction}
\end{figure}

In this section, 
we study how well the encoders of the pretrained models are able to denoise the input. 
For each model, first, we corrupt its inputs using the corresponding noising method 
and then train a linear classifier over its encoder outputs,
using the identity of the original input token as the label, 
similar to \citet{brunner2019identifiability}.
In Figure~\ref{fig:token_prediction} we report the perplexity (\ppl~$\downarrow$) of each classifier evaluated on the wmt18 en\tbi{de} devset.
We report separately the scores for real and corrupted tokens.
We observe that, as expected, 
in all models the \ppl over non-corrupted tokens is almost perfect (\ppl$\approx1$), 
indicating that the encoders perfectly preserve the input in their outputs.
However, the results vary significantly for corrupted tokens.

\shortpar{Masking}
The representations of masked tokens yield the lowest \ppl, 
which implies that token reconstruction happens partially in the encoder.
However, the eMLM loss over the encoder makes the outputs much more predictive of the original tokens.
This shows that the reconstruction loss does not push the encoder to denoise the input well enough, unlike the similar\footnotemark ~but explicit signal from eMLM.

\footnotetext{The eMLM and reconstruction losses are similar, 
but are applied in different places (encoder~vs.~decoder). The signal from reconstruction reaches the encoder through the decoder.}

\shortpar{Shuffling}
The \ppl for original tokens is very low, whereas for shuffled tokens extremely high.
This shows that the encoder does \textit{not} fix the word order but simply relays the input to the output.
This is in line with \citet{Xu2020-vg} who showed that in Transformer-based NMT 
word-reordering happens in the decoder, instead of the encoder.

\shortpar{Replacement}
We observe a huge gap in how predictive the representations of real and replaced tokens are.
We suspect that the encoder is ``misled'' by the replacements and relays their information to its outputs.
Both tying and RTD help the encoder to generate representations that are more predictive of the true input. 
RTD, however, interferes slightly with the representations of real tokens.

\subsection{Decoder Uncertainty}\label{sec:analysis-entropy}

\begin{figure}[t]
\centering
\includegraphics[width=1\columnwidth]{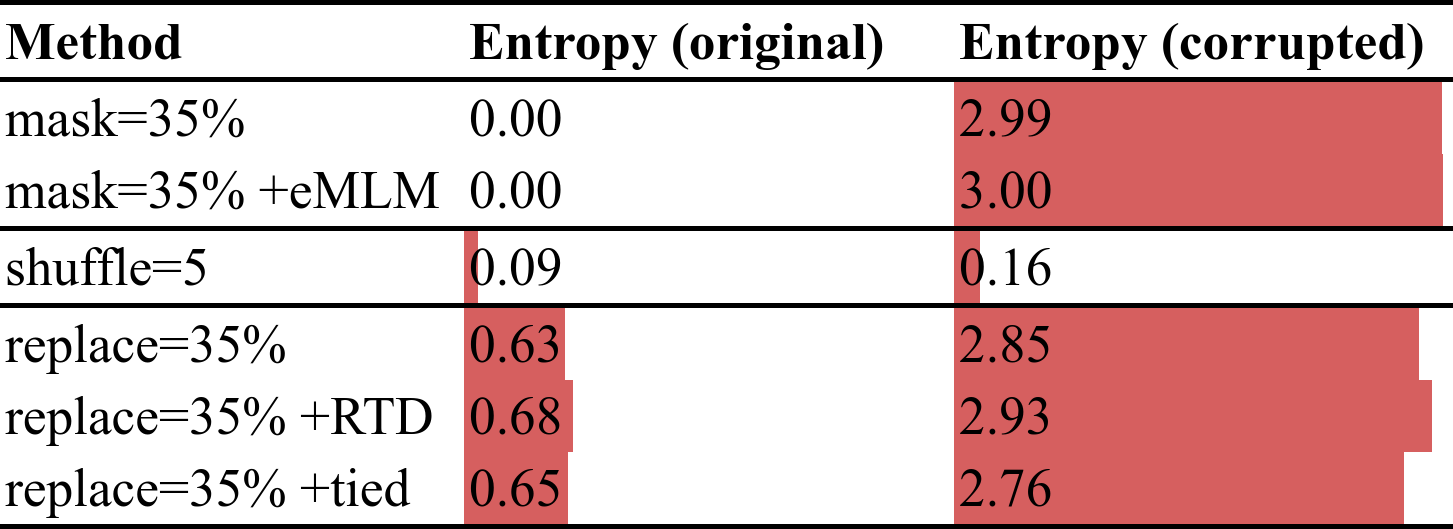}
\caption{Entropy of decoder's distributions during the reconstruction of original and corrupted tokens.}
\label{fig:analysis_entropy}
\end{figure}

Next, we focus on the decoder and study how its token-level uncertainty 
varies while it reconstructs original and corrupted tokens.
For each model, first, we corrupt its inputs using the corresponding noising method 
and then measure the entropy of the decoder's distributions for each target token.
Low entropy values indicate that the decoder predicts the target tokens with certainty,
by exploiting the encoder representations.
Figure~\ref{fig:analysis_entropy} shows the average entropy for original and corrupted tokens.

\shortpar{Masking}
\todo{
There was a bug in the mask+eMLM experiment. 
I fed inputs with replacements instead of masked tokens into. This is why there was so large uncertainty over non-corrupted tokens. I updated the text.
}
When the decoder is presented with masked inputs, it directly copies the unmasked tokens (exactly \textit{zero} entropy).
By contrast, predicting masked tokens is naturally harder, 
and the decoder becomes very uncertain.
Adding eMLM over the encoder does not change this behaviour.

\shortpar{Shuffling}
The decoder of ``shuffle=5'' predicts all tokens with extreme certainty.
Note that, in every step, the decoder has to choose the correct token out of $N$ input tokens, instead of the full vocabulary, and combined with the constraints imposed by the ground-truth prefix, 
it is very easy for the model to find which input word to copy next.
Therefore, we hypothesize that during pretraining the model mainly relies on the LM  capabilities of the decoder.

\shortpar{Replacement}
We observe that the decoder is uncertain not only for fake but to a small extent, for real words as well.
If a replacement is coherent and complies with the grammar of a given language,
the decoder can be misled, which makes it ``question'' the identity of all words.
RTD or tying with the generator show no clear effects.

\subsection{Decoder Sensitivity to Encoder Outputs}\label{sec:decoder-sensitivity}
This analysis aims to estimate the reliance of the decoder on the outputs (i.e., representations) of the encoder. 
First, we feed to the encoder a corrupted sentence
$\bm{x}= \langle x_1, x'_2, x'_3, \ldots, x_N \rangle$, 
where $x'_i$ denotes a corrupted token,
and obtain its outputs $\bm{h}= \langle h_1, h'_2, h'_3, \ldots, h_N \rangle$.
Then, we block the information of $h'_i$ and measure how much it affects the reconstruction loss. 
We consider two blocking methods:
(1) \textit{zeroing}, in which we replace $h'_i$ with a zero vector,
and (2) \textit{mixing}, in which we replace $h'_i$ with random representations from other sentences in a batch.
The amount by which the reconstruction loss increases implies 
how useful is the information in $h'_i$, 
and how sensitive the decoder is to them.

\begin{figure}[t]
    \centering
    \includegraphics[width=1\columnwidth]{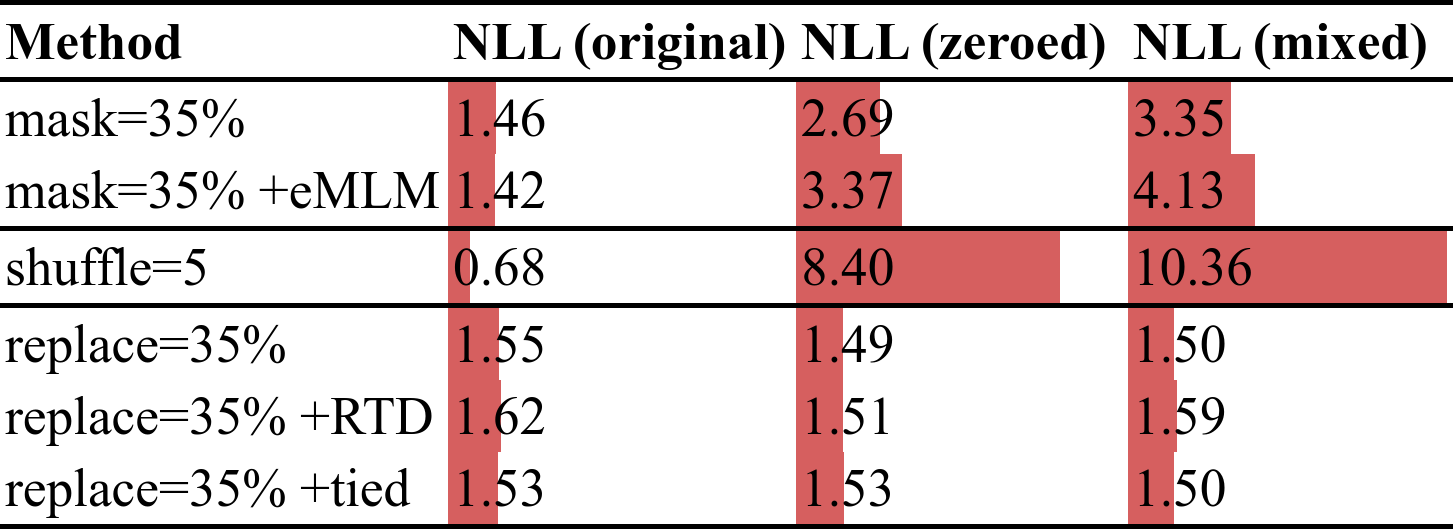}
    \caption{Change in reconstruction loss (NLL $\downarrow$) after blocking the representations of corrupting tokens.}
\label{fig:decoder-sensitivity-deen}
\end{figure}

In Figure~\ref{fig:decoder-sensitivity-deen} we report the differences with and without blocking, in terms of the reconstruction loss (NLL).
The models trained with shuffling originally yield the best reconstruction, 
which shows that it is comparatively the easiest noise for the decoder. 
Replacement is slightly harder than masking noise, 
because with masked inputs the decoder can easily tell when to copy and when to predict,
while replacements can be misleading (recall \S\ref{sec:analysis-entropy}).

\shortpar{Masking} ``mask'' and ``mask+eMLM'' reconstruct the input equally well,
but when the representations of masked tokens are zeroed, ``mask+eMLM'' is affected more.
eMLM forces the reconstruction to partially happen in the encoder (recall \S\ref{sec:analysis-identifiability}), so the decoder relies more on it.
Mixing increases NLL even more, as we inject misleading information.

\shortpar{Replacement} RTD leads to worse reconstruction error, 
which suggests that it is interfering even in the pretraining phase. 
Surprisingly, blocking the representations of replaced words not only does not increase the reconstruction loss 
but even slightly decreases it. 
This implies that during pretraining, the models learn to \textit{ignore} the replaced words.

\shortpar{Shuffling} 
Zeroing the representations of misplaced tokens is destructive and 
replacing them with random representations, 
increases the loss even further.
The decoder focuses so heavily on putting words in the right order and has no ``doubts'' about their identity. 
Therefore, when presented with missing or misleading information, 
instead of ``falling back'' into an unconditional LM, that uses only on the target prefix, it completely  fails\footnotemark.

\footnotetext{NLL after zeroing is 8.4. It equals to \ppl of $\exp(8.4)=4447$,
which is very high even for a weak unconditional LM.}

\subsection{Visualization of Encoder Representations}\label{sec:tsne}
\begin{figure}[t]
    \centering
    \includegraphics[width=1\columnwidth]{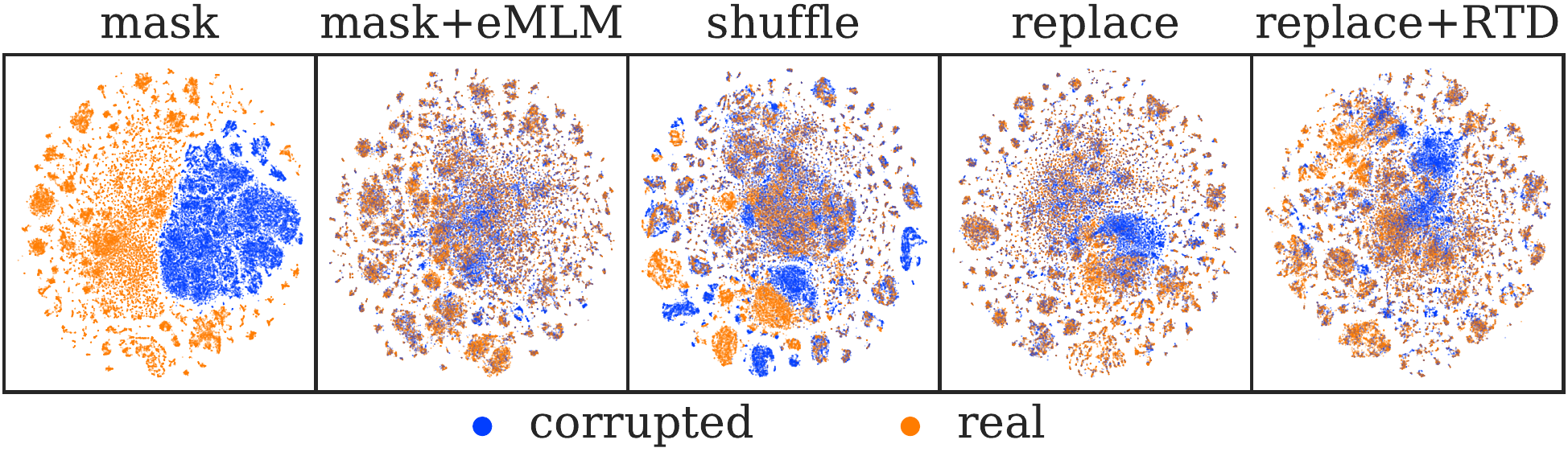}
    \caption{Visualization of encoder representations}
\label{fig:tsne}
\end{figure}

\todo{christos: I removed the visualizations with language information as it doesn't add much and we need the space.}
In Figure~\ref{fig:tsne}, we visualize the encoder token representations using t-SNE~\cite{vanDerMaaten2008} 
and color code them based on whether the belong to original or corrupted tokens
(see Appendix~\ref{sec:viz-tokens} for details and more visualizations).
Masking induces separated representation between masked and unmasked tokens, 
which enables the decoder to copy with certainty (see \S\ref{sec:analysis-entropy}),
while eMLM, that pushes the encoder to reconstruct the corrupted tokens, 
makes them more similar to the original ones (see \S\ref{sec:analysis-identifiability}). 
Although the representations of original and reordered tokens have a large overlap, 
we observe some separated clusters of original and misplaced tokens,
implying that the encoder is partially aware of shuffling noise.
As expected, adding RTD over the ``replace'' models enables allows it to identify more corrupted tokens, 
as shown by the size of the corresponding clusters.

\section{Conclusions}

In this work, we explore new unsupervised pretraining methods for NMT.
We consider alternative objectives to masking, 
such as reordering or replacing input words,
that produce training examples similar to real sentences.

We discover that (semi-)supervised NMT is not very sensitive to the pretraining objective.
While some methods are better than others,
most models converge to similar \bleu scores (\S\ref{sec:sup-nmt}, \S\ref{sec:semi-nmt}).
Surprisingly, even pretraining with shuffled inputs yields competitive results with the other methods.
Our ablation experiments (\S\ref{sec:ablation}) imply that pretraining benefits more the decoder than the encoder.

In unsupervised NMT, however, the results vary significantly (\S\ref{sec:unsup-nmt}).
Shuffling noise leads to significantly worse performance, 
whereas masking noise, unexpectedly, yields the highest \bleu.
Experiments on parallel sentence retrieval (\S\ref{sec:xss}) show that different objectives 
do affect the encoder cross-lingual abilities, 
and are reflected on the unsupervised NMT results.
Through further and extensive analysis of pretrained models (\S\ref{sec:analysis}), 
we find that they encode and use information in different ways.

We conclude that finetuning to each downstream NMT task 
is sensitive to different properties of pretrained models.
(Semi-) Supervised NMT benefits from strong and fluent decoders,
because the signal from the parallel data compensates for encoders with poor cross-lingual representations.
Unsupervised NMT finetuning, however, requires models with good source-target mappings,
and is also sensitive to certain model biases, 
such as the tendency to copy (\S\ref{sec:analysis-entropy}) induced by mask-based pretraining.

\let\thefootnote\relax\footnotetext{
\textbf{Acknowledgments}
This work was conducted within the EU project Gourmet, under grant agreement No 825299. It was also supported by the UK EPSRC grant EP/S001271/1 (MTStretch). It was performed using computing resources in CSD3 using funding from the UK EPSRC (capital grant EP/P020259/1), and DiRAC funding from the Science and Technology Facilities Council.}

\bibliography{refs_used}
\bibliographystyle{acl_natbib}

\clearpage
\newpage
\appendix

\section{Model Configuration \& Training}\label{sec:model-config}
We use 6 Transformer layers in both the encoder and the decoder, with embedding/hidden size of 512, feed-forward filter size of 2048, 8 attention heads and we apply 0.1 dropout to all layers.
We optimize our models using Adam~\cite{kingma2014Adam} with $\beta_1=0.9$, $\beta_1=0.999$, and $\epsilon=10^{-6}$. All models use sinusoidal positional embeddings.

We tie the weights of the embedding and output (projection) layers of  all\footnote{we assume a shared multi-lingual vocabulary}
sub-networks~\cite{E17-2025, Inan2017TyingWV}, which involves the encoder, decoder and MLM generator.
For pretraining, we use a learning rate of $5e^{-4}$ with a linear warm-up of 16K steps, followed by inverted squared decay. We train each model for 300K steps with mini-batches of 24K tokens on 8 Nvidia V100 GPUs, which requires approximately 4-5 days.
The maximum sentence length is set to 256 tokens.
For the finetuning experiments, we use a learning rate of $3e^{-5}$ with a linear warm-up of 2.5K steps and mini-batches of 12K tokens. We finetune each model for 60K in the supervised setting and 120K steps in the semi-supervised setting. 
We also use increased the dropout to 0.3 and set label smoothing~\cite{szegedy2016rethinking} to 0.1,
to avoid over-fitting on the limited parallel data.

\section{Unsupervised NMT}\label{sec:appendix-unmt}
Instead of adding input noise, like ~\citet{artetxe2018unsupervised, lample2018unsupervised},
we follow the finetuning process of mBART~\cite{Liu2020-un}.
To prevent models from copying the source during backtranslation 
and force the transition to the translation task,
we allow only the most frequent tokens in the target language\footnotemark to be generated for the first 2K steps.
Specifically, we mask tokens with frequency less than $10^{-3}$, as measure in the monolingual data.
For model selection, we use a small subset of 200 sentences from the wmt18 devset.

\subsection{Supervised Translation Ablations}\label{sec:appendix-ablation}
\begin{figure}[t]
    \centering
    \includegraphics[width=0.9\columnwidth,page=5]{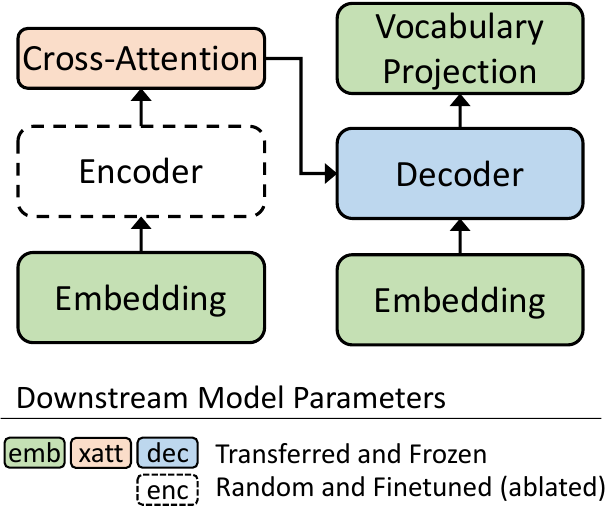}
    \caption{
    Visualization of the ablation experiment, using the ablation of the encoder as example.
    The figure shows the high-level architecture of a model and the colors correspond to the parameter set.
    The input embeddings of the encoder, the decoder and the vocabulary projection 
    have the same color as they share the same parameters.
    During finetuning, we update only the weights of the ablated component, and all the other (pretrained) weights are frozen.
    }
    \label{fig:ablation-protocol}
\end{figure}

In Figure~\ref{fig:ablation-protocol}, we visualize the experimental protocol for our ablation experiment.
To test (i.e., ablate) a component, the process is the following:

\begin{enumerate}
    \item We transfer all the weights of a pretrained model to a downstream model, 
except the weights of the ablated component.
    \item We freeze the pretrained weights and finetune only the ablated (i.e., randomly initialized) component.
\end{enumerate}

We decided to follow this protocol, in order to isolate the effects on the final \bleu score on the ablated component, 
and to also prevent the other components from compensating. In concurrent work, \citet{Gheini2021OnTS} have considered a similar experimental protocol, but to study a different but related phenomenon.
In Figure~\ref{fig:ablation-protocol}, we show the ablation results for the en\tto{de} direction.

\begin{figure}[t]
    \centering
    \includegraphics[width=1\columnwidth]{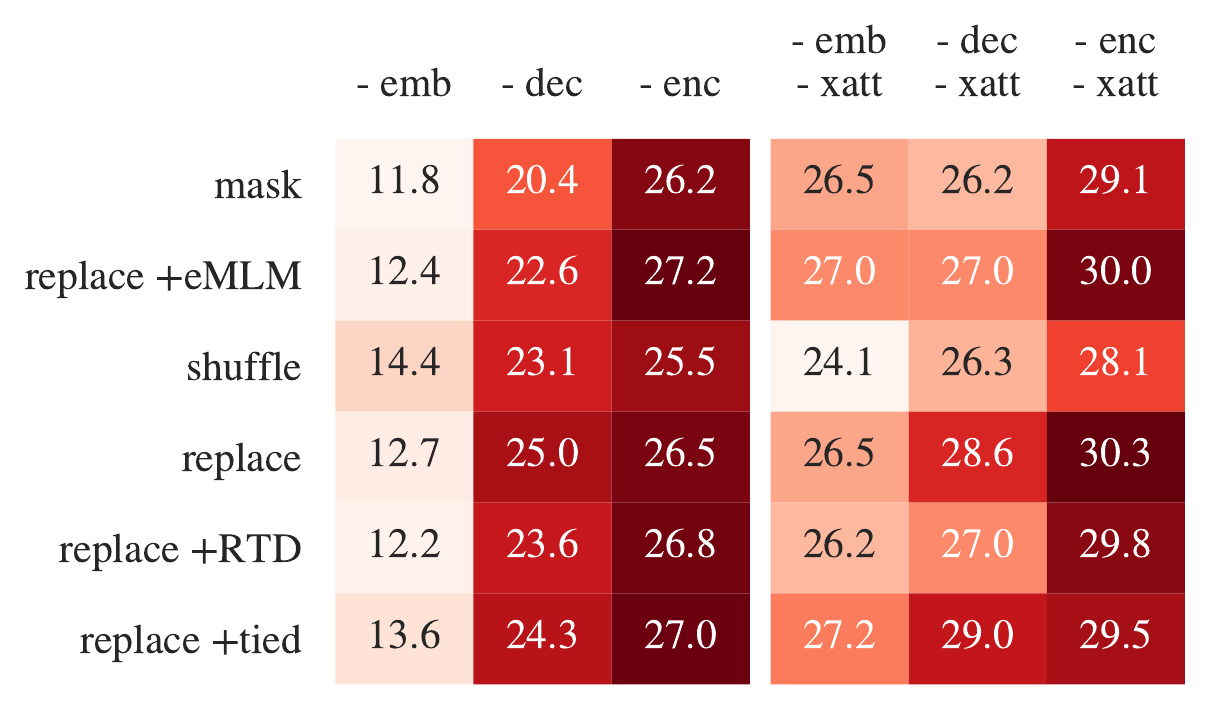}
    \caption{
    Ablation results for supervised NMT on en\tto{de} (wmt19). 
    We reset each main component individually (left) and with the cross-attention (right).
    }
    \label{fig:ablation-ende-merged-frozen}
\end{figure}

\section{Analysis Results in Other Languages}\label{sec:appendix-analysis}
\begin{figure}[H]
\centering
\includegraphics[width=1\linewidth]{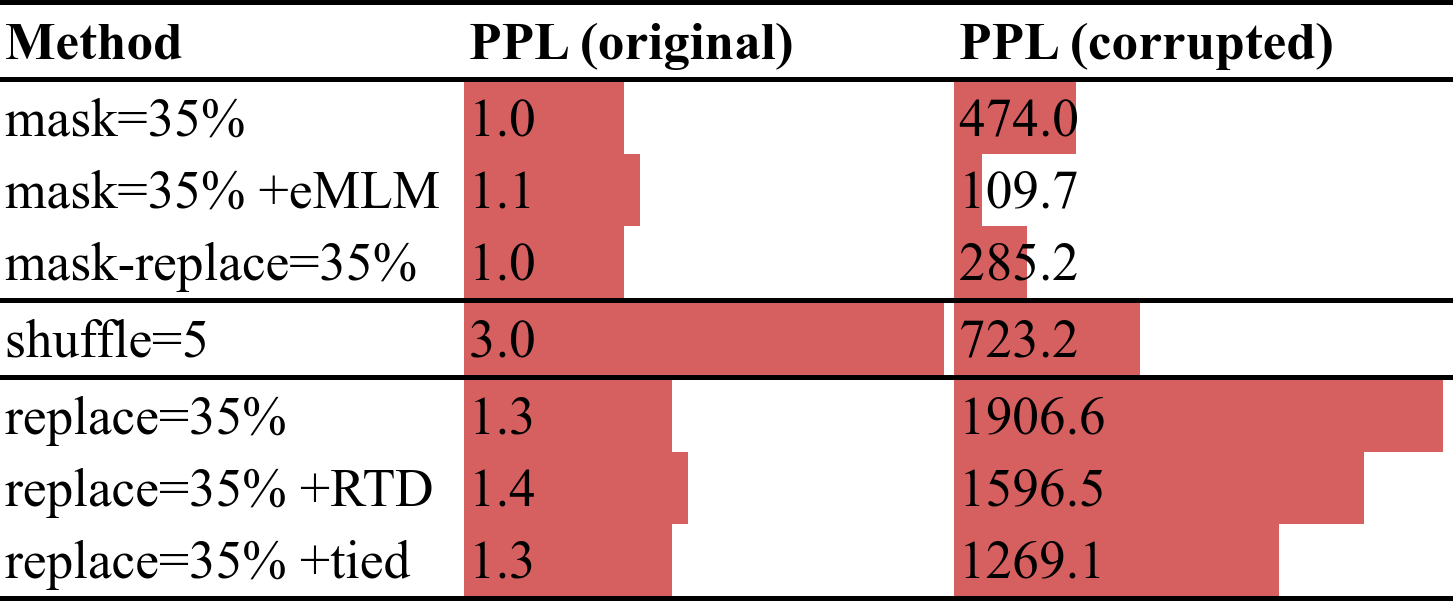}
\caption{Perplexity (PPL$\downarrow$) of the token prediction probe for the en\tbi{ne} pretrained models.}
\label{fig:analysis_token_prediction_sien}
\end{figure}

\begin{figure}[H]
\centering
\includegraphics[width=1\linewidth]{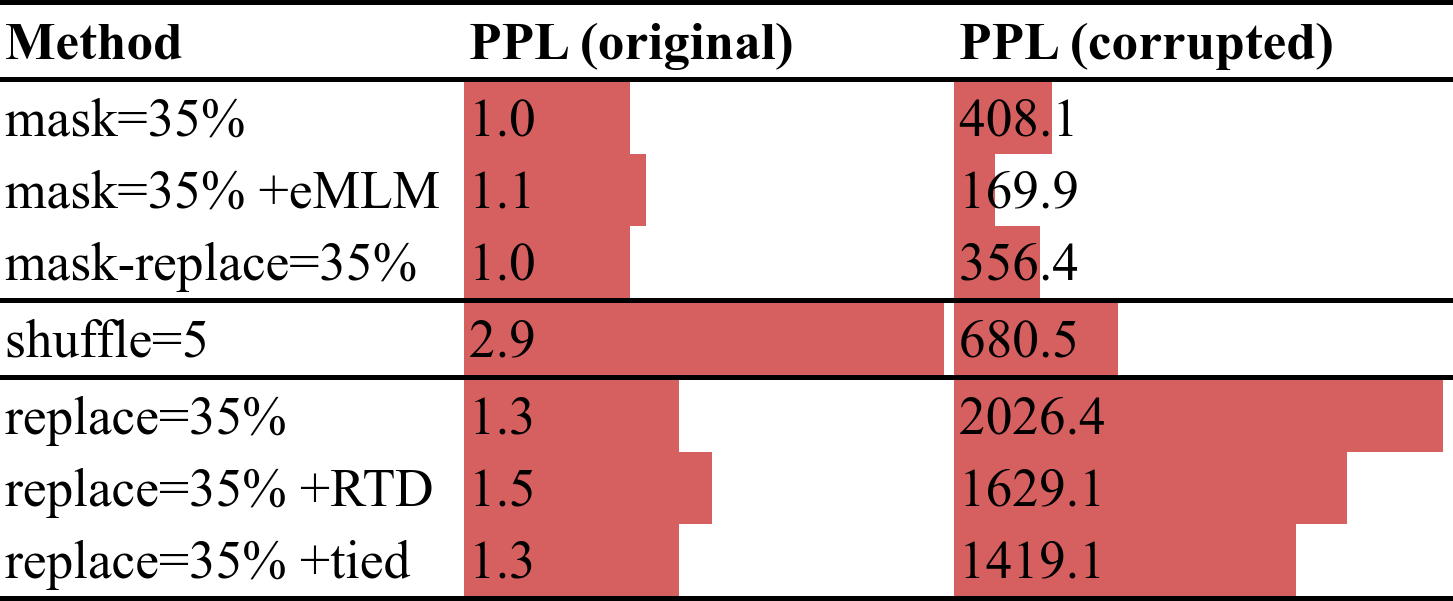}
\caption{Perplexity (PPL$\downarrow$) of the token prediction probe for the en\tbi{si} pretrained models.}
\label{fig:analysis_token_prediction_neen}
\vspace{-15pt}
\end{figure}

\begin{figure}[H]
\centering
\includegraphics[width=1\linewidth]{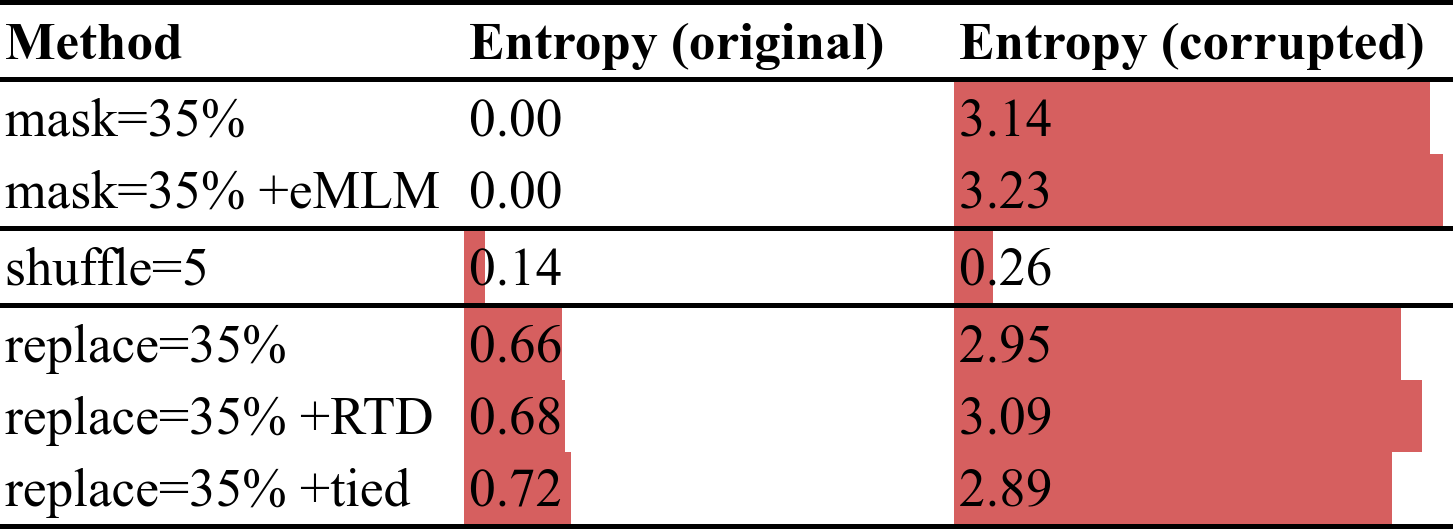}
\caption{Decoder entropy for the reconstruction of real/corrupted tokens, for the en\tbi{si} pretrained models.}
\label{fig:analysis_entropy_sien}
\vspace{-15pt}
\end{figure}

\begin{figure}[H]
\centering
\includegraphics[width=1\linewidth]{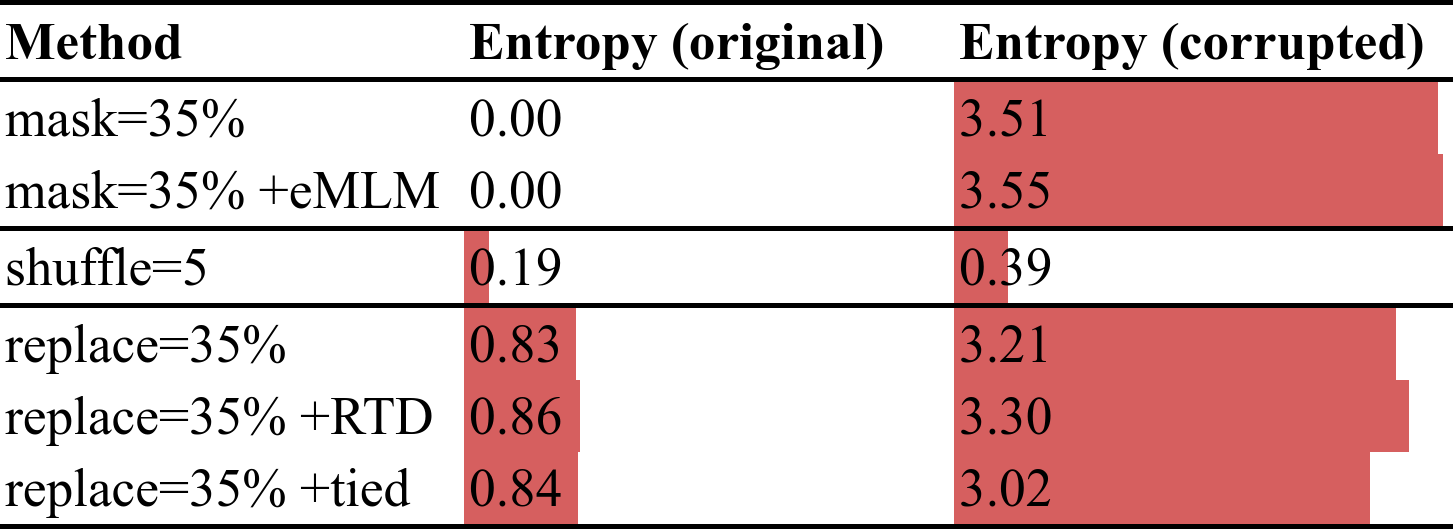}
\caption{Decoder entropy for the reconstruction of real/corrupted tokens, for the en\tbi{ne} pretrained models.}
\label{fig:analysis_entropy_neen}
\vspace{-15pt}
\end{figure}

\begin{figure}[H]
\centering
\includegraphics[width=1\linewidth]{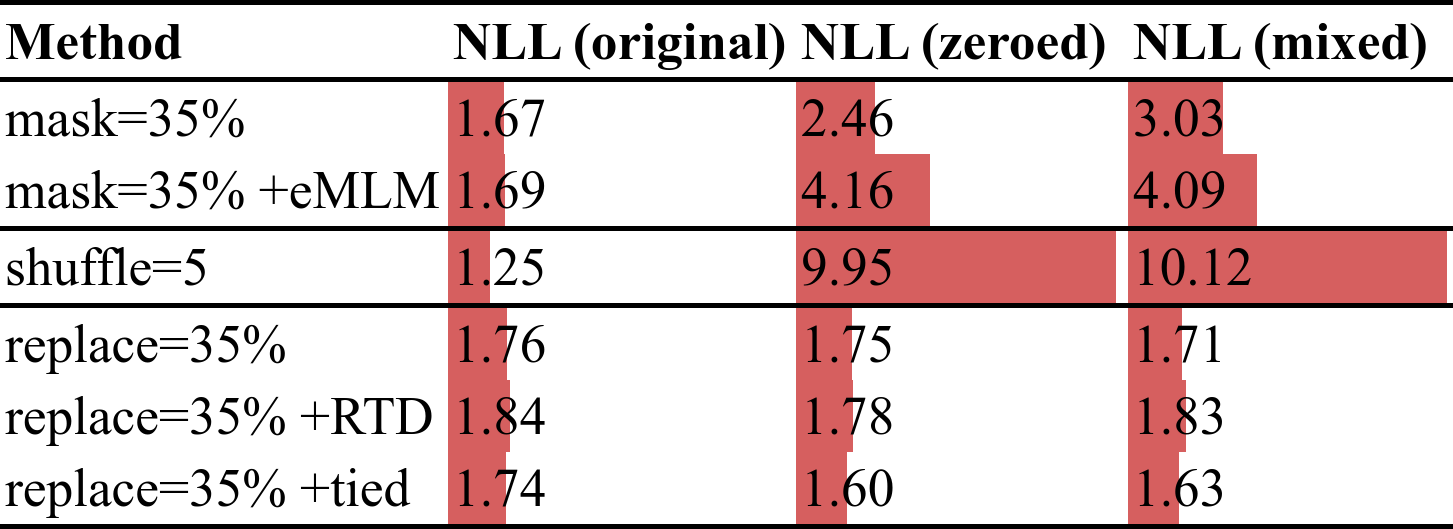}
    \caption{Reconstruction loss (NLL $\downarrow$) with/without blocking the outputs of corrupted tokens (en\tbi{si}).
    }\label{fig:analysis_decoder_sensitivity_nll_sien}
\vspace{-15pt}
\end{figure}

\begin{figure}[H]
\centering
\includegraphics[width=1\linewidth]{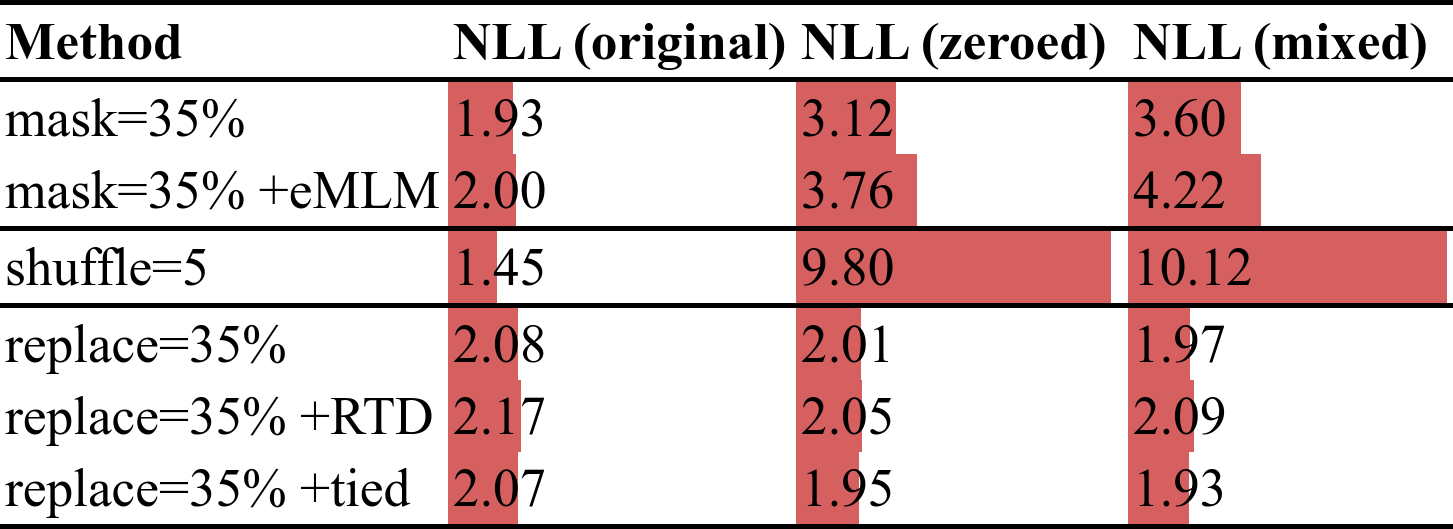}
    \caption{Reconstruction loss (NLL $\downarrow$) with/without blocking the outputs of corrupted tokens (en\tbi{ne}).
    }\label{fig:analysis_decoder_sensitivity_nll_neen}
\end{figure}

In Figures
~\ref{fig:analysis_token_prediction_sien},
~\ref{fig:analysis_token_prediction_neen},
~\ref{fig:analysis_entropy_sien},
~\ref{fig:analysis_entropy_neen},
~\ref{fig:analysis_decoder_sensitivity_nll_sien},
~\ref{fig:analysis_decoder_sensitivity_nll_neen}, we report the analysis results~\ref{sec:analysis} on the en\tbi{ne} and \tbi{ne} pretrained models, evaluated on their NMT dev sets.
We observe that the results are very consistent with those for en\tbi{de}.

\paragraph{Training Details of Probing Classifier}
For the analysis in Sec.~\ref{sec:analysis-identifiability}, each classifier is trained on the monolingual data for 50K steps,
and optimized with Adam using a learning rate of 0.0001.
Only the parameters of the classifier are updated and the rest of the model remains \textit{fixed}.


\subsection{\hspace{-5pt}Visualization of Encoder Representations}\label{sec:viz-tokens}
In this section, we visually inspect the input token representations.
We compare each method based on how the representations evolve through the layers of the encoder,
by focusing on two aspects of each token, (1) its \textit{language} and (2) whether it has been \textit{corrupted} or not. 
The goal is to inspect how each model encodes these two types of information about the input tokens.

\paragraph{Methodology}
We sample 5K sentences from the English-German monolingual data and pass them through the encoder of each model using corresponding noising method.
For each token, we keep its representations from every layer and label them by language and noise.
We keep only the representations of the 2K most frequent tokens 
and exclude the representations of special tokens, such as the \texttt{[BOS]}, \texttt{[EOS]}, and language IDs, which significantly skew the results. 
The final dataset contains approximately 100K token representations per layer (600K in total).
For the visualization, 
we project to 2D with t-SNE~\cite{vanDerMaaten2008}.
For each model, we visualize the representations of its encoder per layer (L1 to L6), which we colour-code by language (top-row) and the identity (bottom-row) of each token.
L1 refers to the outputs of the first Transformer layer, therefore the tokens have been contextualized once.

\begin{figure}[H]
    \centering
    \includegraphics[width=1\linewidth]{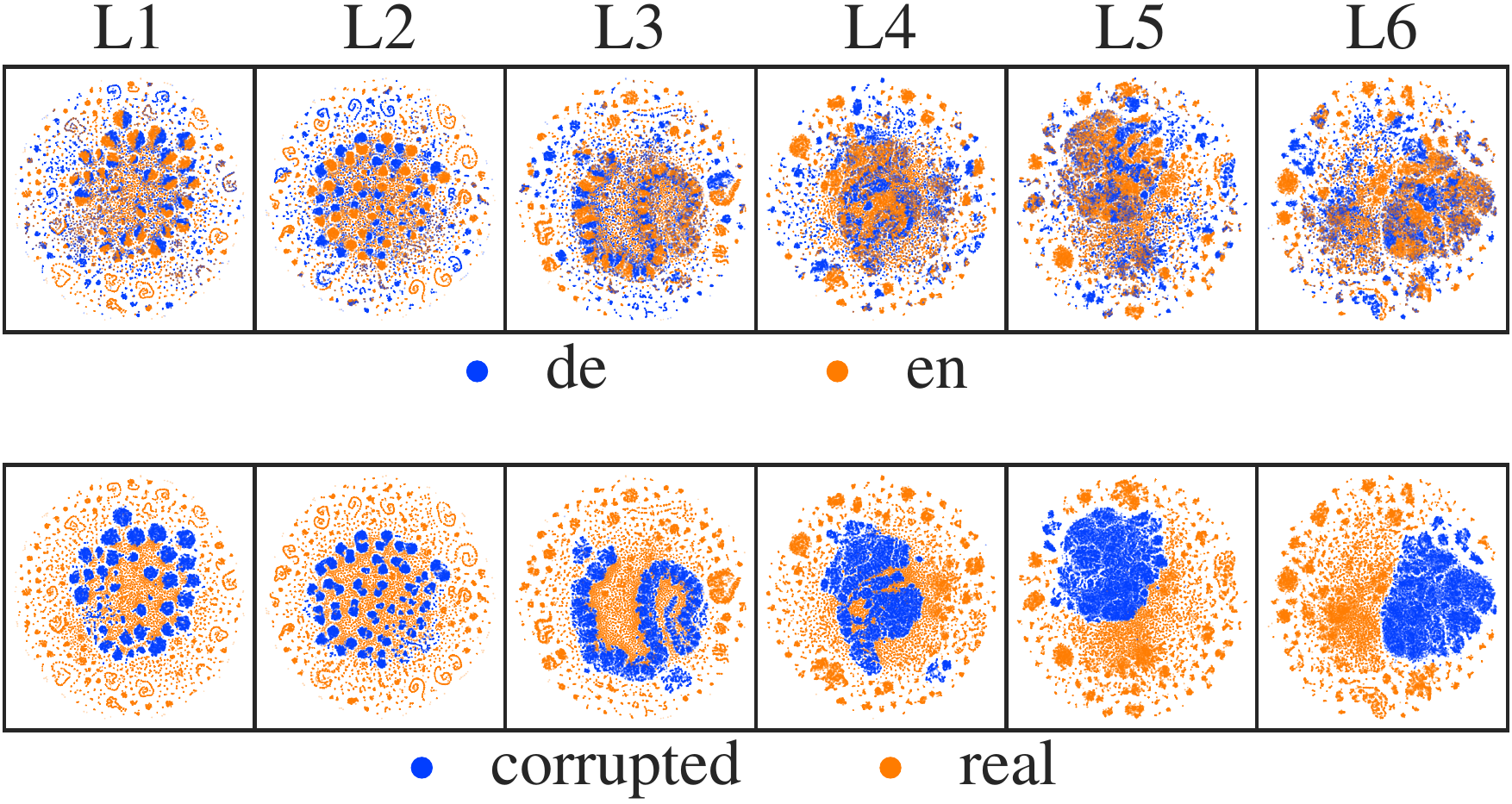}
    \caption{Visualization of encoder representations from the ``mask=35\%'' model.}
\label{fig:tokens-mask35}
\end{figure}

\paragraph{Masking}
In Fig.~\ref{fig:tokens-mask35} we visualize the encoder of the ``mask=35\%'' pretrained model.
Real and masked tokens occupy different regions,
whereas the tokens from each language are much closer to each other.
All models trained with masking noise exhibit similar behaviour.
In the first layer, the masked tokens are organized into multiple small clusters,
but the encoder progressively groups them into larger structures.
We visually verify that the encoder keeps the masked tokens separated even in the last layer.
This aligns with our findings in Sec.~\ref{sec:analysis-identifiability}, which suggest that
the reconstruction loss does not incentivize the encoder to denoise the representations of masked tokens.
Also, it enables the model to easily identify the real tokens and can copy them, as discussed in Sec.~\ref{sec:analysis-entropy}.

\begin{figure}[H]
    \centering
    \includegraphics[width=1\linewidth]{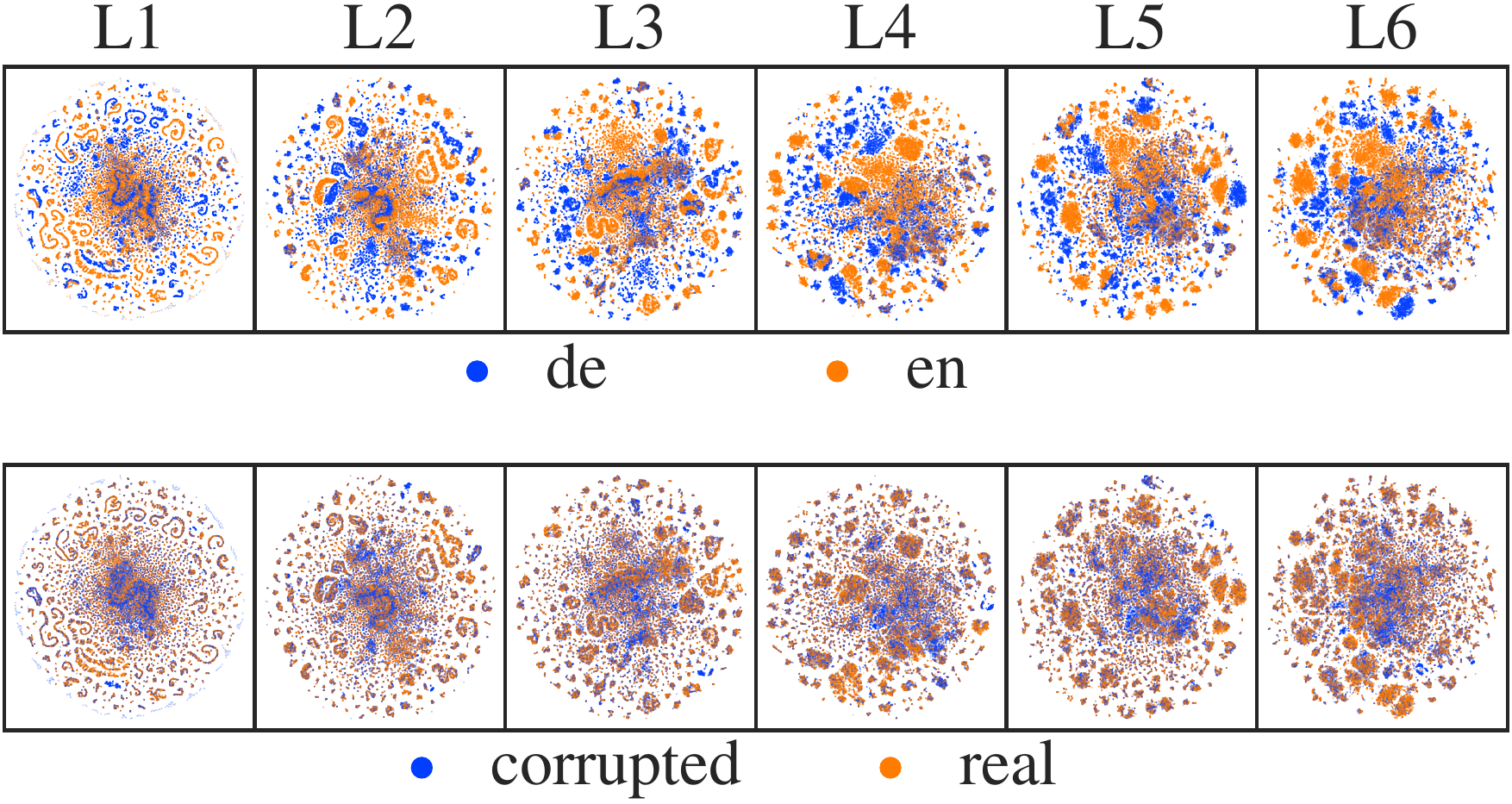}
    \caption{Visualization of encoder representations from the ``mask=35\%+eMLM'' model.}
    \label{fig:tokens-mask35+mlm}
\end{figure}
\paragraph{Masking+eMLM}
In Fig.~\ref{fig:tokens-mask35+mlm} we visualize the ``mask=35\%+MLM'' model.
Adding eMLM makes the masked and real tokens are indistinguishable from each other.
Intuitively, to minimize the MLM loss the encoder must generate representations that are predictive of the true identity of the masked tokens, therefore similar to real tokens.
We observe a small overlap between English and German tokens, as there are more language-specific clusters.
We believe that is because the eMLM loss pushes the representation to better encode the grammar and semantics of each token.

\begin{figure}[H]
    \centering
    \includegraphics[width=1\linewidth]{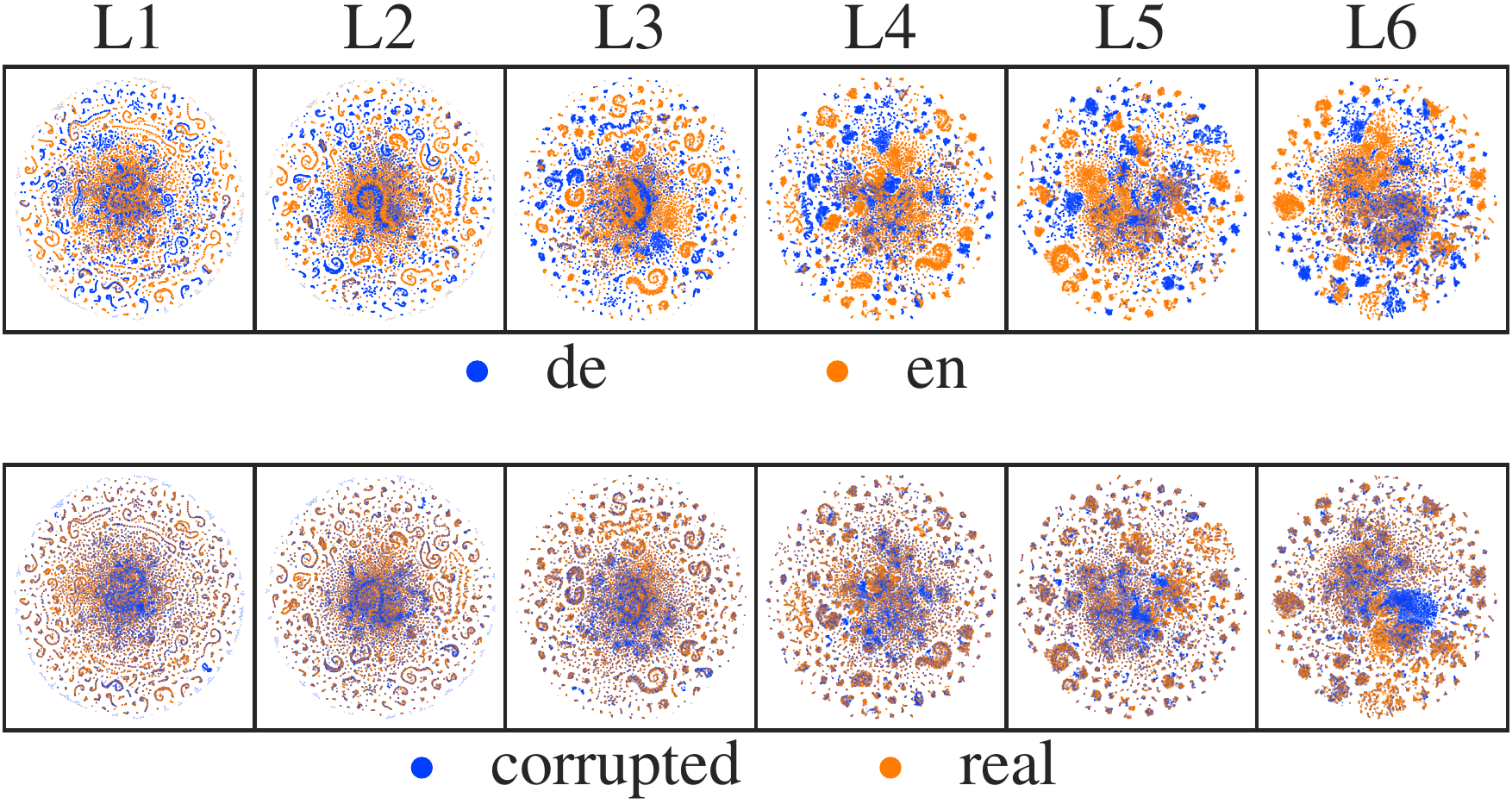}
    \caption{Layer-wise visualization of encoder representations from the ``replace=35\%'' model.}
    \label{fig:tokens-replace35}
\end{figure}
\paragraph{Replacement}
In Fig.~\ref{fig:tokens-replace35} we visualize the ``replace=35\%'' model.
There is a moderate separation between languages, slightly less than the ``mask=35\%+MLM'' model. 
However, we observe that the representations of real and fake tokens generally overlap with each other, unlike the ``mask=35\%'' model, especially in the lower layers.
This is because the model is always given as input embeddings of actual words and not \texttt{[MASK]}, which unless contextualized all of them are treated the same.
Only in the last layer, we can see the formation of a fake-only cluster.
This suggests that the pretraining objective (i.e., reconstruction) creates a bias towards discriminating between real and fake token.
However, the separation is not as extreme as in the masked-based models but is not obvious if the encoder can't or is not biased to more clearly separate them.

\begingroup
\setlength{\tabcolsep}{8.0pt} %
\renewcommand{\arraystretch}{1.0} %
\begin{table*}[t]
	\small
	\centering
	\begin{tabular}{lrrrrrrrr}
		\toprule
		\textbf{Method} & \multicolumn{2}{c}{\trans{en}{de}} & \multicolumn{2}{c}{\trans{de}{en}}
		& \trans{en}{ne} & \trans{ne}{en} & \trans{en}{si} & \trans{si}{en} \\ \cmidrule(lr){2-3} \cmidrule(lr){4-5}
		                              & \multicolumn{1}{r}{wmt18} & \multicolumn{1}{r}{wmt19} & wmt18 & wmt19 &     &      &     &      \\ \midrule
		random                      & \cell{26.2}{0.1}   & \cell{25.3}{0.1} & \cell{27.6}{0.1}  & \cell{19.1}{0.3}  		& \cell{3.3}{0.1} & \cell{6.5}{0.1} 		& \cell{2.5}{0.1} & \cell{6.5}{0.1} \\
		mask=35\%                     & \cell{33.3}{0.1}   & \cell{30.7}{0.2} & \cell{33.2}{0.0}  & \best{25.4}{0.0}  		& \cell{5.1}{0.1} & \cell{10.2}{0.1} 		& \cell{3.7}{0.0} & \cell{10.0}{0.1} \\
		mask=35\% +eMLM                & \cell{33.4}{0.0}   & \cell{30.6}{0.1} & \cell{33.5}{0.1}  & \cell{25.2}{0.2}  		& \best{5.3}{0.0} & \best{10.8}{0.1} 		& \best{4.0}{0.0} & \cell{10.4}{0.1} \\
		mask=35\% (span)              & \cell{33.3}{0.1}   & \cell{30.5}{0.1} & \cell{33.4}{0.0}  & \cell{25.2}{0.0}  		& \cell{5.1}{0.1} & \cell{10.1}{0.1} 		& \cell{3.9}{0.1} & \cell{9.9}{0.1} \\
		shuffle=5                     & \cell{31.6}{0.1}   & \cell{28.7}{0.0} & \cell{31.7}{0.0}  & \cell{23.9}{0.1}  		& \cell{4.9}{0.0} & \cell{9.9}{0.1} 		& \cell{3.4}{0.0} & \cell{10.1}{0.1} \\
		replace=35\%                  & \cell{33.9}{0.0}   & \cell{30.3}{0.2} & \cell{33.5}{0.1}  & \best{25.4}{0.1}  		& \cell{5.1}{0.1} & \cell{9.9}{0.1} 		& \cell{3.7}{0.0} & \cell{9.8}{0.0} \\
		replace=35\% +RTD             & \cell{32.9}{0.1}   & \cell{30.0}{0.0} & \cell{32.5}{0.0}  & \cell{24.4}{0.1}  		& \cell{5.0}{0.0} & \cell{9.9}{0.1} 		& \cell{3.4}{0.1} & \cell{9.7}{0.2} \\
		replace=35\% +tied            & \best{34.2}{0.0}   & \best{30.8}{0.1} & \best{33.7}{0.1}  & \cell{25.3}{0.2}  		& \best{5.3}{0.0} & \cell{10.6}{0.1} 		& \cell{3.7}{0.0} & \best{10.5}{0.1} \\ \cmidrule{2-9}
		\quad\quad+ shuffle=3         & \cell{34.0}{0.0}   & \gain{31.1}{0.1} & \cell{33.4}{0.1}  & \cell{25.1}{0.2}  		& \gain{5.5}{0.0} & \gain{11.0}{0.0} 		& \cell{4.0}{0.0} & \gain{10.8}{0.1} \\
		\quad\quad+ dec:~mask=15\%    & \cell{33.9}{0.1}   & \gain{30.9}{0.0} & \cell{33.6}{0.1}  & \cell{25.3}{0.2}  		& \gain{5.5}{0.0} & \cell{10.5}{0.0} 		& \cell{3.9}{0.0} & \cell{10.4}{0.0} \\
		\quad\quad+ dec:~replace=15\% & \gain{34.5}{0.1}   & \cell{30.7}{0.1} & \cell{33.4}{0.0}  & \gain{25.6}{0.1}  		& \gain{5.6}{0.0} & \cell{10.5}{0.1} 		& \cell{3.9}{0.0} & \gain{10.7}{0.1} \\
		\bottomrule
	\end{tabular}
	\caption{Finetuning results to supervised NMT.
	    ``dec:X'' denotes method that add noise to the the decoder.
		We report the average of 3 runs and the standard error of the mean. 
    }
	\label{table:sup_nmt_dec}
\end{table*}
\endgroup

\begin{figure}[H]
    \centering
    \includegraphics[width=1\linewidth]{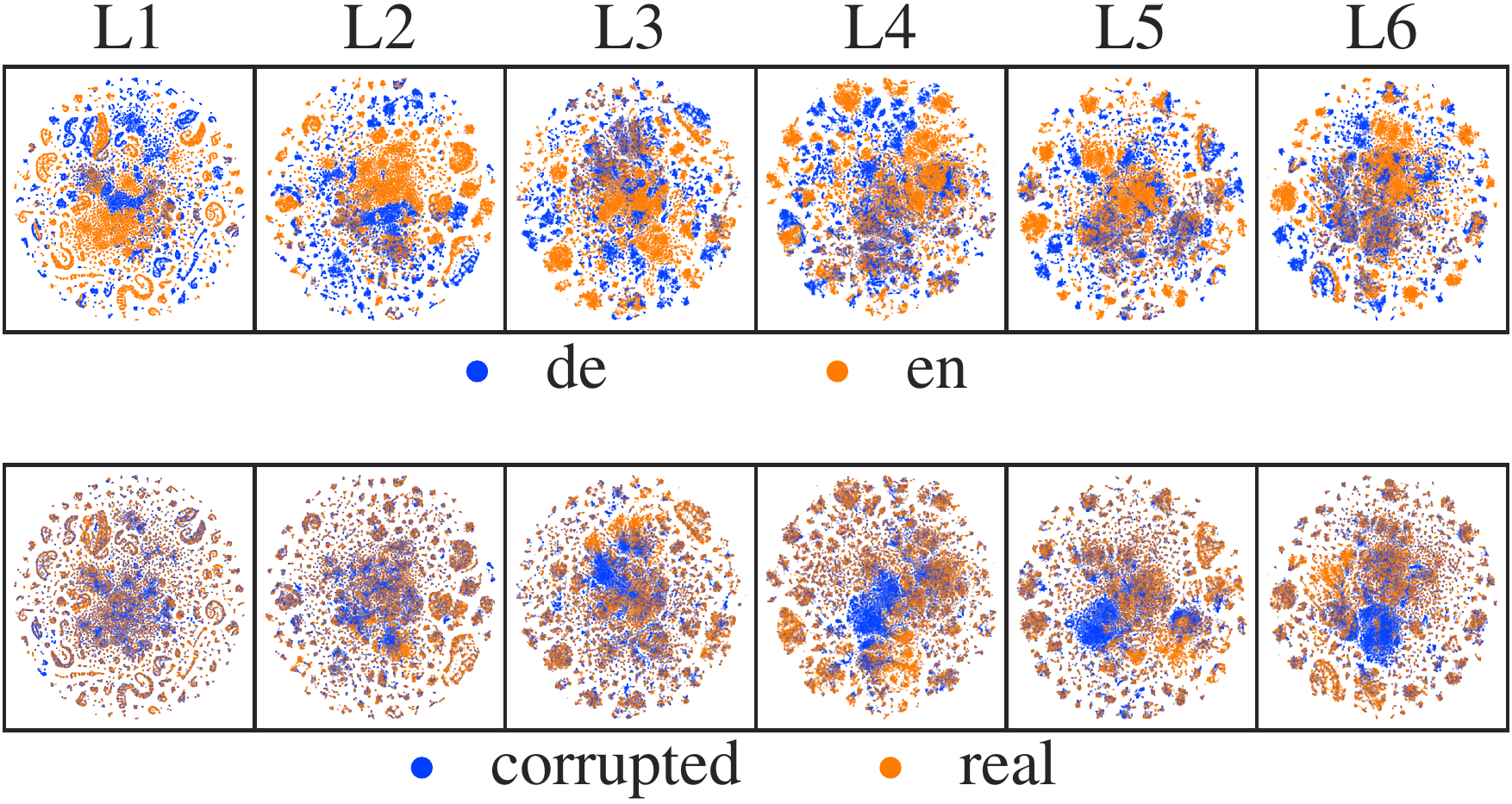}
    \caption{Visualization of representations from the ``replace=35\%+RTD=4'' model, that is trained with an RTD head over the 4th layer of the encoder.}
    \label{fig:tokens-replace35+rtd}
\end{figure}
\paragraph{Replacement Token Detection}
In Fig.~\ref{fig:tokens-replace35+rtd} we visualize the ``replace=35\%+RTD=4'' model, that is trained with an RTD head over the 4th layer of the encoder.
The only difference with the ``replace=35\%+RTD=6'' is that the effects of RTD start to show earlier.
Compared to the ``replace=35'' the representations are more clustered and less spread-out, even in the lower layers.
The real and fake tokens are much better separated, and the separation peaks at layer 4, which visually verifies the bias introduced by RTD,
but the separation is not as extreme as for masking noise.
Although this suggests that separating masked/original words is harder than real/fake, 
it also depends on the weight used for the RTD loss during pretraining, 
which is not something we have explored.
Also, the visualization suggests that the separation remains approximately constant in the remaining layers.
There is no perceptible difference with ``replace=35'' in terms of language.

\begin{figure}[H]
    \centering
    \includegraphics[width=1\linewidth]{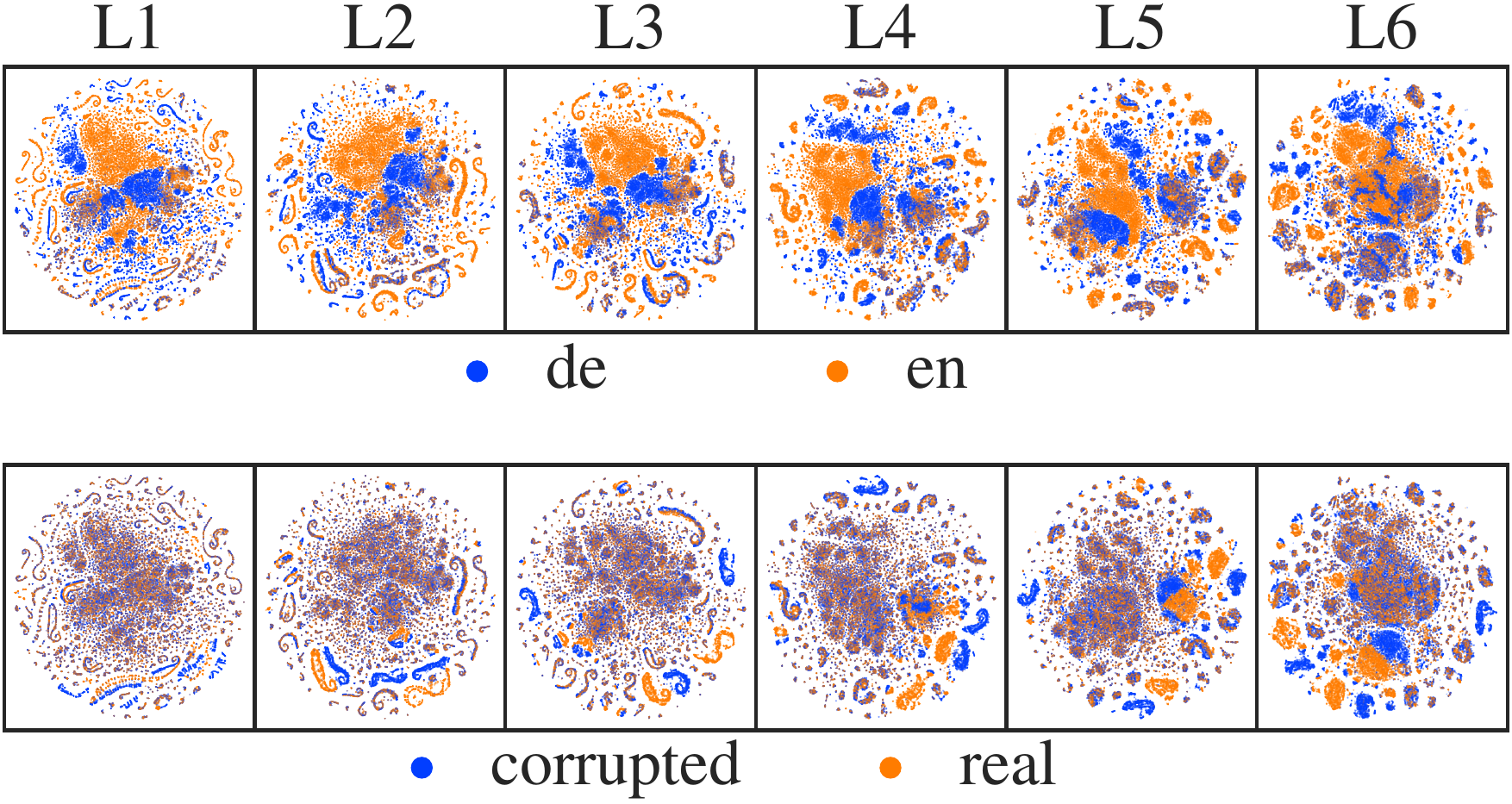}
    \caption{Visualization of representations from the ``shuffle=5'' model.}\label{fig:tokens-shuffle}
\end{figure}
\paragraph{Shuffling}
In Fig.~\ref{fig:tokens-shuffle} we visualize the ``shuffle=5'' model.
We observe a strong separation between languages, that slightly decreases in the upper layers.
The language clusters are relatively large, unlike the much smaller and local language-specific clusters seen in the other models.
As for noise, initially all tokens are represented similarly, but as the encoder re-contextualizes the input, 
it puts more misplaced tokens in separate clusters.
This shows that the model becomes progressively more ``aware'' about the misplaced tokens in the input.
After manual inspection of the clusters with corrupted tokens, we found that the majority of them contain punctuation marks. 
This makes intuitive sense, as it should be easy for the model to identify punctuation marks that has been misplaced.

\section{Additional NMT Experiments}\label{sec:decoder-noise}

In Table~\ref{table:sup_nmt_dec} we report some additional results for supervised NMT~\ref{sec:sup-nmt} omitted from the main paper.
Besides combining shuffling and word replacements in the input, 
we also introduce noise in the decoder side,
by randomly replacing 15\% words of the decoder's input with the \texttt{[MASK]} token (``dec:mask=15\%'') or with samples from the generator (``dec:replace=15\%'').
Note that, we reduce the amount of noise in this case to avoid disrupting training\footnote{\citet{bowman-etal-2016-generating} reported that masking more than 20\% of words in the decoder hurts its LM capabilities.}.
Overall, we observe that both methods increase performance in some language pairs,
but the improvements are marginal.

We also experimented with RTD over shuffled inputs, by training the model to explicitly detect if words were misplaced or not, but this configuration lead to poor results.

\end{document}